\documentclass{article}

\usepackage[preprint]{corl_2021}

\usepackage[utf8]{inputenc} %
\usepackage[T1]{fontenc}    %
\usepackage{url}            %
\usepackage{booktabs}       %
\usepackage{amsfonts}       %
\usepackage{nicefrac}       %
\usepackage{microtype}      %

\usepackage{microtype}
\usepackage{graphicx}
\usepackage{subfigure}
\usepackage[utf8]{inputenc} %
\usepackage[T1]{fontenc}    %
\usepackage{url}            %
\usepackage{booktabs}       %
\usepackage{amsfonts}       %
\usepackage{nicefrac}       %
\usepackage{color}
\usepackage{mathtools}
\usepackage{amsmath,amssymb}
\usepackage[svgnames]{xcolor}
\usepackage{bm}
\usepackage{siunitx}
\usepackage{wrapfig}
\usepackage{algorithm}
\usepackage[noend]{algorithmic}
\usepackage{lipsum}

\usepackage{amsmath,amsfonts,bm}

\def\eqref#1{equation~\ref{#1}}

\def\1{\bm{1}}

\def\va{{\bm{a}}}

\DeclareMathAlphabet{\mathsfit}{\encodingdefault}{\sfdefault}{m}{sl}
\SetMathAlphabet{\mathsfit}{bold}{\encodingdefault}{\sfdefault}{bx}{n}

\DeclarePairedDelimiterX{\norm}[1]{\lVert}{\rVert}{#1}

\newcommand{\ie}{i.e., }
\newcommand{\eg}{e.g., }
\newcommand{\Skip}[1]{}

\newcommand{\skild}[0]{SkiLD}

\title{Demonstration-Guided Reinforcement Learning \\with Learned Skills}

\author{%
  Karl Pertsch\thanks{Correspondence to \href{mailto:pertsch@usc.edu}{\texttt{pertsch@usc.edu}}}, \;\;Youngwoon Lee, \;\;Yue Wu, \;\;Joseph J. Lim \\[0.2cm]
  University of Southern California \\
  \href{https://clvrai.com/skild}{\url{https://clvrai.com/skild}} 
}

\begin{document}

\maketitle

\begin{abstract}
Demonstration-guided reinforcement learning (RL) is a promising approach for learning complex behaviors by leveraging both reward feedback and a set of target task demonstrations. Prior approaches for demonstration-guided RL treat every new task as an independent learning problem and attempt to follow the provided demonstrations step-by-step, akin to a human trying to imitate a completely unseen behavior by following the demonstrator's exact muscle movements.
Naturally, such learning will be slow, but often new behaviors are not completely unseen: they share subtasks with behaviors we have previously learned. In this work, we aim to exploit this shared subtask structure to increase the efficiency of demonstration-guided RL. 
We first learn a set of reusable skills from large offline datasets of prior experience collected across many tasks. We then propose \textbf{Ski}ll-based \textbf{L}earning with \textbf{D}emonstrations (\textbf{SkiLD}), an algorithm for demonstration-guided RL that efficiently leverages the provided demonstrations by following the demonstrated \emph{skills} instead of the primitive actions, resulting in substantial performance improvements over prior demonstration-guided RL approaches.
We validate the effectiveness of our approach on long-horizon maze navigation and complex robot manipulation tasks.

\end{abstract}

\section{Introduction}
\label{sec:intro}

\begin{wrapfigure}{r}{0.5\textwidth}
    \centering
    \vspace{-0.5cm}
    \includegraphics[width=1\linewidth]{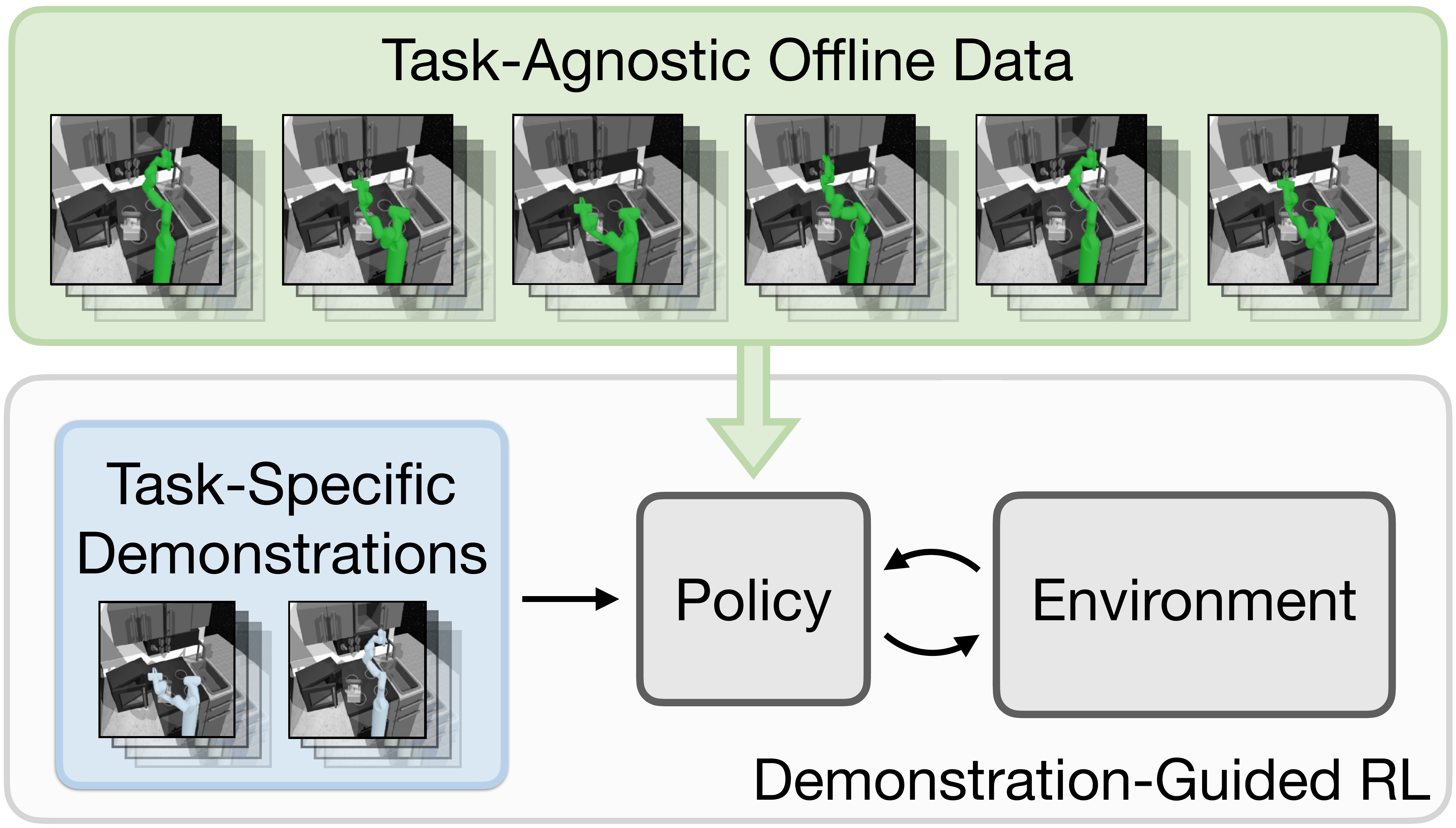}
    \vspace{-0.5cm}
    \caption{
    Our approach \skild~leverages large, task-agnostic datasets collected across many different tasks for efficient demonstration-guided reinforcement learning by (1)~acquiring a rich motor skill repertoire from such offline data and (2)~understanding and imitating the demonstrations based on the skill repertoire. 
    }
    \label{fig:overview}
    \vspace{-0.5cm}
\end{wrapfigure}

Humans are remarkably efficient at acquiring new skills from demonstrations: often a single demonstration of the desired behavior and a few trials of the task are sufficient to master it~\citep{bekkering2000imitation, al2001specificity, hodges2007modelled}. To allow for such efficient learning, we can leverage a large number of previously learned behaviors~\citep{al2001specificity, hodges2007modelled}. Instead of imitating precisely each of the demonstrated muscle movements, humans can extract the performed \emph{skills} and use the rich repertoire of already acquired skills to efficiently reproduce the desired behavior.

Demonstrations are also commonly used in reinforcement learning (RL) to guide exploration and improve sample efficiency~\citep{vecerik2017leveraging, hester2018deep, rajeswaran2018learning,nair2018overcoming,zhu2018reinforcement}. However, such demonstration-guided RL approaches attempt to learn tasks \emph{from scratch}: analogous to a human trying to imitate a completely unseen behavior by following every demonstrated muscle movement, they try to imitate the \emph{primitive actions} performed in the provided demonstrations. As with humans, such step-by-step imitation leads to brittle policies~\cite{ross2011dagger}, and thus these approaches require many demonstrations and environment interactions to learn a new task.

We propose to improve the efficiency of demonstration-guided RL by leveraging prior experience in the form of an offline ``task-agnostic`` experience dataset, collected not on one but across many tasks (see Figure~\ref{fig:overview}). Given such a dataset, our approach extracts reusable skills: robust  short-horizon behaviors that can be recombined to learn new tasks. Like a human imitating complex behaviors via the chaining of known skills, we can use this repertoire of skills for efficient demonstration-guided RL on a new task by guiding the policy using the demonstrated \emph{skills} instead of the primitive actions. %

Concretely, we propose \textbf{Ski}ll-based \textbf{L}earning with \textbf{D}emonstrations (\textbf{SkiLD}), a demonstration-guided RL algorithm that %
learns short-horizon skills from offline datasets and then learns new tasks efficiently by leveraging these skills to follow a given set of demonstrations. Across challenging navigation and robotic manipulation tasks our approach significantly improves the learning efficiency over prior demonstration-guided RL approaches.

In summary, the contributions of our work are threefold: (1)~we introduce the problem of leveraging task-agnostic offline datasets for accelerating demonstration-guided RL on unseen tasks, (2)~we propose \skild, a skill-based algorithm for efficient demonstration-guided RL and (3)~we show the effectiveness of our approach on a maze navigation and two complex robotic manipulation tasks. %

\section{Related Work}
\label{sec:related_work}

\textbf{Imitation learning.} Learning from Demonstration, also known as imitation learning ~\cite{argall2009survey}, is a common approach for learning complex behaviors by leveraging a set of demonstrations. Most prior approaches for imitation learning are either based on behavioral cloning (BC, ~\citep{pomerleau1989alvinn}), which uses supervised learning to mimic the demonstrated actions, or inverse reinforcement learning~(IRL, \citep{abbeel2004apprenticeship,ho2016generative}), which infers a reward from the demonstrations and then trains a policy to optimize it. However, BC commonly suffers from distribution shift and struggles to learn robust policies~\cite{ross2011dagger}, while IRL's joint optimization of reward and policy can result in unstable training. %

\textbf{Demonstration-guided RL.} A number of prior works aim to %
mitigate these problems by combining reinforcement learning with imitation learning. This allows the agent to leverage demonstrations for overcoming exploration challenges in RL while using RL to increase robustness and performance of the imitation learning policies. Prior work on demonstration-guided RL can be categorized into three groups: (1)~approaches that use BC to initialize and regularize policies during RL training~\cite{rajeswaran2018learning,nair2018overcoming}, (2)~approaches that place the demonstrations in the replay buffer of an off-policy RL algorithm~\cite{vecerik2017leveraging,hester2018deep}, and (3)~approaches that augment the environment rewards with rewards extracted from the demonstrations~\cite{zhu2018reinforcement,peng2018deepmimic, merel2017learning}. While these approaches improve the efficiency of RL, they treat each new task as an \emph{independent} learning problem, \ie attempt to learn policies without taking any prior experience into account. As a result, they require many demonstrations to learn effectively, which is especially expensive since a new set of demonstrations needs to be collected for every new task. %

\textbf{Online RL with offline datasets.} As an alternative to expensive task-specific demonstrations, multiple recent works have proposed to accelerate reinforcement learning by leveraging \emph{task-agnostic} experience in the form of large datasets collected across many tasks~\cite{pertsch2020spirl,siegel2020keep,nair2020accelerating,ajay2020opal,singh2020parrot,singh2020cog}. In contrast to demonstrations, such task-agnostic datasets can be collected cheaply from a variety of sources like autonomous exploration~\citep{hausman2018learning,sharma2019dynamics} or human tele-operation~\citep{gupta2019relay,mandlekar2018roboturk,lynch2020learning}, %
but will lead to slower learning than demonstrations since the data is not specific to the downstream task.

\textbf{Skill-based RL.} One class of approaches for leveraging such offline datasets that is particularly suited for learning long-horizon behaviors is skill-based RL~\cite{hausman2018learning,merel2018neural,kipf2018compositional,merel2019reusable,shankar2019discovering,whitney2019dynamics,gupta2019relay,lee2020learning,lynch2020learning,pertsch2020keyin,pertsch2020spirl}. These methods extract reusable skills from task-agnostic datasets and learn new tasks by recombining them. Yet, such approaches perform \emph{reinforcement learning} over the set of extracted skills to learn the downstream task. Although being more efficient than RL over primitive actions, they still require many environment interactions to learn long-horizon tasks. In our work we combine the best of both worlds: by using large, task-agnostic datasets and a small number of task-specific demonstrations, we accelerate the learning of long-horizon tasks while reducing the number of required demonstrations.

\section{Approach}
\label{sec:approach}

\begin{figure*}[t]
    \centering
    \includegraphics[width=1\linewidth]{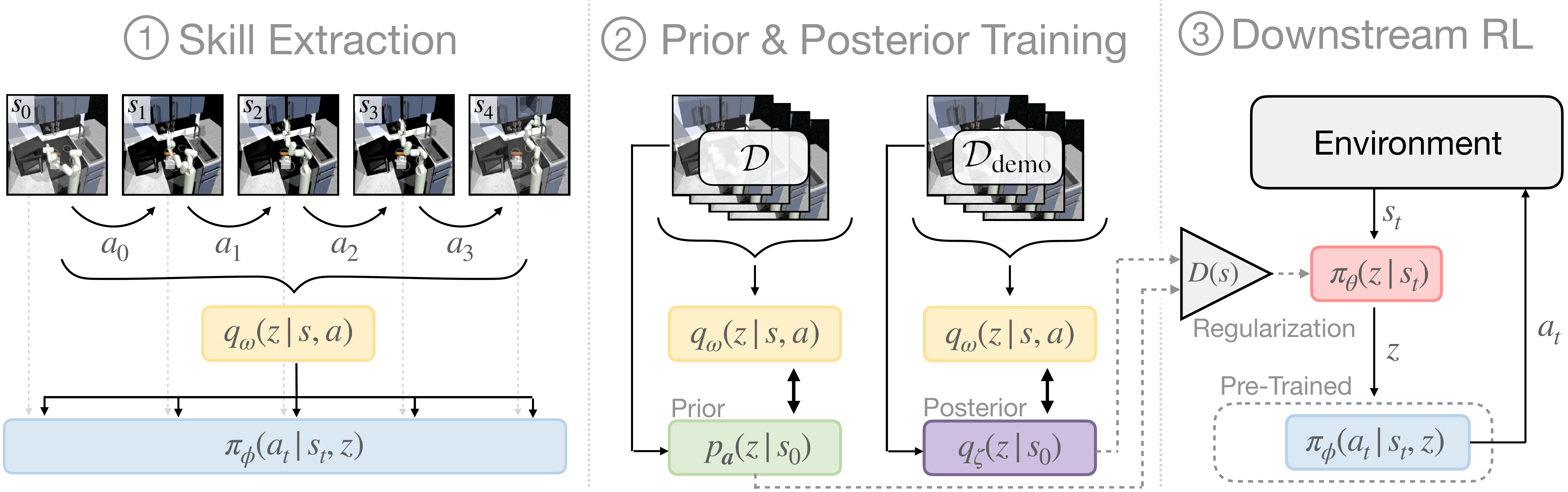}
    \vspace{-0.5cm}
    \caption{Our approach, \skild, combines task-agnostic experience and task-specific demonstrations to efficiently learn target tasks in three steps: (1)~extract skill representation from task-agnostic offline data, (2)~learn task-agnostic skill prior from task-agnostic data and task-specific skill posterior from demonstrations, and (3)~learn a high-level skill policy for the target task using prior knowledge from both task-agnostic offline data and task-specific demonstrations. \textbf{Left}: Skill embedding model with skill extractor (\textcolor[HTML]{F4C430}{\textbf{yellow}}) and closed-loop skill policy (\textcolor[HTML]{0F52BA}{\textbf{blue}}). \textbf{Middle}: Training of skill prior (\textcolor[HTML]{00A86B}{\textbf{green}}) from task-agnostic data and skill posterior (\textcolor[HTML]{8D79AA}{\textbf{purple}}) from demonstrations. \textbf{Right}: Training of high-level skill policy (\textcolor[HTML]{FF9090}{\textbf{red}}) on a downstream task using the pre-trained skill representation and regularization via the skill prior and posterior, mediated by the demonstration discriminator $D(s)$.
    }
    \label{fig:model}
\end{figure*}

Our goal is to use skills extracted from task-agnostic prior experience data to improve the efficiency of demonstration-guided RL on a new task. We aim to leverage a set of provided demonstrations by following the performed \emph{skills} as opposed to the primitive actions. Therefore, we need a model that can (1)~leverage prior data to learn a rich set of skills and (2)~identify the skills performed in the demonstrations in order to follow them. Next, we formally define our problem, summarize relevant prior work on RL with learned skills and then describe our demonstration-guided RL approach.

\subsection{Preliminaries}
\label{sec:prelim}

\paragraph{Problem Formulation} 
We assume access to two types of datasets: a large task-agnostic offline dataset and a small task-specific demonstration dataset. The task-agnostic dataset $\mathcal{D} = \{s_t, a_t, ...\}$ consists of trajectories of meaningful agent behaviors, but includes no demonstrations of the target task. We only assume that its trajectories contain \emph{short-horizon} behaviors that can be reused to solve the target task. Such data can be collected without a particular task in mind using a mix of sources, \eg via human teleoperation, autonomous exploration, or through policies trained for other tasks. Since it can be used to accelerate \emph{many} downstream task that utilize similar short-term behaviors we call it \emph{task-agnostic}. In contrast, the task-specific data is a much smaller set of demonstration trajectories $\mathcal{D}_{\text{demo}} = \{s^d_t, a^d_t, ...\}$ that are specific to a single target task.

The downstream learning problem is formulated as a Markov decision process (MDP) defined by a tuple $(\mathcal{S}, \mathcal{A}, \mathcal{T}, R, \rho, \gamma)$ of states, actions, transition probabilities, rewards, initial state distribution, and discount factor. We aim to learn a policy $\pi_\theta(a \vert s)$ with parameters $\theta$ that maximizes the discounted sum of rewards $J(\theta) = \mathbb{E}_\pi \big[ \sum_{t=0}^{T-1} J_t \big] = \mathbb{E}_\pi \big[ \sum_{t=0}^{T-1} \gamma^t r_t \big]$, where $T$ is the episode horizon.

\paragraph{Skill Prior RL}
Our goal is to extract skills from task-agnostic experience data and reuse them for \emph{demonstration-guided} RL. Prior work has investigated the reuse of learned skills for accelerating RL~\cite{pertsch2020spirl}. %
In this section, we will briefly summarize their proposed approach Skill Prior RL (SPiRL) and then describe how our approach improves upon it in the \emph{demonstration-guided} RL setting. %

SPiRL defines a skill as a sequence of $H$ consecutive actions $\va = \{a_t, \dots, a_{t+H-1}\}$, where the skill horizon $H$ is a hyperparameter. It uses the task-agnostic data to jointly learn (1)~a generative model of skills $p(\va \vert z)$, that decodes latent skill embeddings $z$ into executable action sequences $\va$, and (2)~a state-conditioned prior distribution $p(z \vert s)$ over skill embeddings. For learning a new downstream task, SPiRL trains a high-level skill policy $\pi(z \vert s)$ whose outputs get decoded into executable actions using the pre-trained skill decoder. Crucially, the learned skill prior is used to guide the policy during downstream RL by maximizing the following divergence-regularized RL objective:
\begin{equation}
\label{eq:spirl_objective}
    J(\theta) = \mathbb{E}_{\pi_\theta} \bigg[\sum_{t=0}^{T-1} r(s_t, z_t) - \alpha D_{\text{KL}}\big(\pi_\theta(z_t \vert s_t), p_\va(z_t \vert s_t)\big)\bigg].
\end{equation}
Here, the KL-divergence term ensures that the policy remains close to the learned skill prior, guiding exploration during RL. By combining this guided exploration with temporal abstraction via the learned skills, SPiRL substantially improves the efficiency of RL on long-horizon tasks.

\subsection{Skill Representation Learning}
\label{sec:subgoal_model}

We leverage SPiRL's skill embedding model for learning our skill representation. We follow prior work on skill-based RL~\cite{lynch2020learning,ajay2020opal} and increase the expressiveness of the skill representation by replacing SPiRL's low-level skill decoder $p(\va \vert z)$ with a closed-loop skill policy $\pi(a \vert s, z)$ that is conditioned on the current environment state. In our experiments we found this closed-loop decoder to improve performance (see Section~\ref{sec:skill_rep_comparison} for an empirical comparison).

Figure~\ref{fig:model} (left) summarizes our skill learning model. It consists of two parts: the skill inference network $q_\omega(z \vert s_{0:H-1}, a_{0:H-2})$ and the closed-loop skill policy $\pi_\phi(a_t \vert s_t, z_t)$. Note that in contrast to SPiRL the skill inference network is state-conditioned to account for the state-conditioned low-level policy. During training we randomly sample an $H$-step state-action trajectory from the task-agnostic dataset and pass it to the skill inference network, which predicts the low-dimensional skill embedding $z$. This skill embedding is then input into the low-level policy $\pi_\phi(a_t \vert s_t, z)$ for every input state. The policy is trained to imitate the given action sequence, thereby learning to reproduce the behaviors encoded by the skill embedding $z$. 

The latent skill representation is optimized using variational inference, which leads to the full skill learning objective:
\begin{equation}
    \max_{\phi, \omega} \mathbb{E}_q \bigg[\underbrace{\prod_{t=0}^{H-2} \log \pi_\phi(a_t \vert s_t, z)}_{\text{behavioral cloning}}
    - \beta \big(\underbrace{\log q_\omega(z \vert s_{0:H-1}, a_{0:H-2})- \log p(z)}_{\text{embedding regularization}}\big) \bigg].
\end{equation}
We use a unit Gaussian prior $p(z)$ and weight the embedding regularization term with a factor $\beta$~\cite{higgins2017beta}.

\subsection{Demonstration-Guided RL with Learned Skills}

\begin{wrapfigure}{r}{0.4\textwidth}
    \centering
    \vspace{-1cm}
    \includegraphics[width=0.9\linewidth]{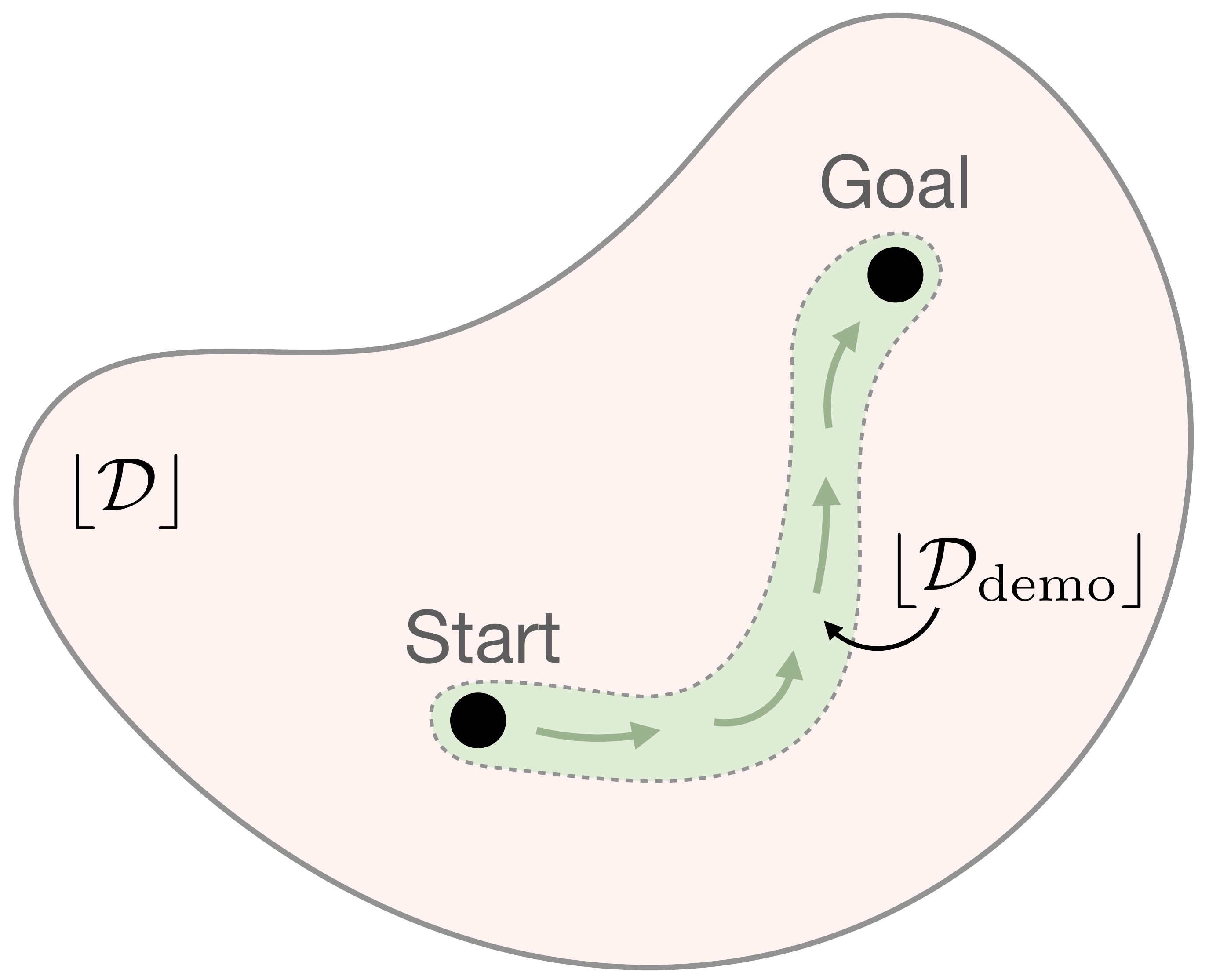}
    \caption{We leverage prior experience data $\mathcal{D}$ and demonstration data $D_\text{demo}$. Our policy is guided by the task-specific skill posterior $q_\zeta(z \vert s)$ within the support of the demonstrations (\textcolor[HTML]{00A86B}{\textbf{green}}) and by the task-agnostic skill prior $p_\va(z \vert s)$ otherwise (\textcolor[HTML]{FF9090}{\textbf{red}}). The agent also receives a reward bonus for reaching states in the demonstration support.
    }
    \vspace{-0.5cm}
    \label{fig:pp_intuition}
\end{wrapfigure}

To leverage the learned skills for accelerating demonstration-guided RL on a new task, we use a hierarchical policy learning scheme (see Figure~\ref{fig:model}, right): a high-level policy $\pi_\theta(z \vert s)$ outputs latent skill embeddings $z$ that get decoded into actions using the pre-trained low-level skill policy. We freeze the weights of the skill policy during downstream training for simplicity.\footnote{Joint optimization of high-level and low-level policy is possible, but we did not find it necessary in our experiments. Prior work on hierarchical RL~\citep{levy2017learning} suggests that joint optimization can lead to training instabilities. We will leave an investigation of this for future work.} %

Our goal is to leverage the task-specific demonstrations to guide learning of the high-level policy on the new task. In Section~\ref{sec:prelim}, we showed how SPiRL~\cite{pertsch2020spirl} leverages a learned \emph{skill prior} $p_\va(z|s)$ to guide exploration. However, this prior is task-agnostic, \ie it encourages exploration of \emph{all} skills that are meaningful to be explored, independent of which task the agent is trying to solve. %
Even though SPiRL's objective makes learning with a large number of skills more efficient, it encourages the policy to explore many skills that are not relevant to the downstream task.

In this work, we propose to extend the skill prior guided approach and leverage target task demonstrations to additionally learn a \emph{task-specific} skill distribution, which we call \emph{skill posterior}~$q_\zeta(z \vert s)$ (in contrast to the skill prior it is conditioned on the target task, hence ``posterior''). We train this skill posterior by using the pre-trained skill inference model $q_\omega(z \vert s_{0:H-1}, a_{0:H-2})$ to extract the embeddings for the skills performed in the demonstration sequences (see Figure~\ref{fig:model}, middle):
\begin{equation}
    \min_\zeta \mathbb{E}_{(s,a) \sim \mathcal{D}_\text{demo}} D_{\text{KL}}\big( q_\omega(z \vert s_{0:H-1}, a_{0:H-2}), q_\zeta(z \vert s_0) \big),
\end{equation}
where $D_{\text{KL}}$ denotes the Kullback-Leibler divergence.

A naive approach for leveraging the skill posterior is to simply use it to replace the skill prior in Equation~\ref{eq:spirl_objective}, \ie to regularize the policy to stay close to the skill posterior in every state. However, the trained skill posterior is only accurate within the support of the demonstration dataset $\lfloor\mathcal{D}_\text{demo}\rfloor$. Since $\vert \mathcal{D}_\text{demo} \vert \ll \vert \mathcal{D} \vert$, this support will only be a small subset of all states (see Figure~\ref{fig:pp_intuition}) and thus the skill posterior will often provide incorrect guidance in states outside the demonstrations' support.

Instead, we propose to use a three-part objective to guide the policy during downstream learning. Our goal in formulating this objective is to (1)~follow the skill posterior \emph{within} the support of the demonstrations, (2)~follow the skill prior \emph{outside} the demonstration support, and (3)~encourage the policy to reach states \emph{within} the demonstration support. Crucial for all three components is to determine whether a given state is within the support of the demonstration data. We propose to use a learned discriminator $D(s)$ to answer this question. $D(s)$ is a binary classifier that distinguishes demonstration and non-demonstration states and it is trained using samples from the demonstration and task-agnostic datasets, respectively. Once trained, we use its output to weight terms in our objective that regularize the policy towards the skill prior or posterior. Additionally, we provide a reward bonus for reaching states which the discriminator classifies as being within the demonstration support. This results in the following term $J_t$ for \skild's full RL objective:
\begin{equation*}
    J_t = \tilde{r}(s_t, z_t) 
    \underbrace{- \alpha_q D_\text{KL}(\pi_\theta(z_t \vert s_t), q_\zeta(z_t \vert s_t)) \cdot D(s_t)}_{\text{posterior regularization}}
    \underbrace{- \alpha D_\text{KL}(\pi_\theta(z_t \vert s_t), p_\va(z_t \vert s_t)) \cdot (1-D(s_t))}_{\text{prior regularization}},
\end{equation*}
\vspace{-10pt}
\begin{equation}
\label{eq:skil_objective}
    \quad\quad\text{with } \tilde{r}(s_t, z_t) =\; (1-\kappa) \cdot r(s_t, z_t) 
    \underbrace{+ \;\kappa \cdot \big[\log D(s_t) - \log\big(1 - D(s_t)\big)\big]}_{\text{discriminator reward}}.
\end{equation}
The weighting factor $\kappa$ is a hyperparameter; $\alpha$ and $\alpha_q$ are either constant or tuned automatically via dual gradient descent~\cite{haarnoja2018sac_algo_applications}. 
The discriminator reward follows common formulations used in adversarial imitation learning~\cite{finn2016connection,fu2018learning,zhu2018reinforcement,kostrikov2018discriminatoractorcritic}.\footnote{We found that using the pre-trained discriminator weights led to stable training, but it is possible to perform full adversarial training by finetuning $D(s)$ with rollouts from the downstream task training. We report results for initial experiments with discriminator finetuning in Section~\ref{sec:skill_imitation} and leave further investigation for future work.} Our formulation combines IRL-like and BC-like objectives by using learned rewards \emph{and} trying to match the demonstration's skill distribution.

For policy optimization, we use a modified version of the SPiRL algorithm~\cite{pertsch2020spirl}, which itself is based on Soft Actor-Critic~\cite{haarnoja2018sac}. Concretely, we replace the environment reward with the discriminator-augmented reward and all prior divergence terms with our new, weighted prior-posterior-divergence terms from equation~\ref{eq:skil_objective} (for the full algorithm see appendix, Section~\ref{sec:algorithm}).

\section{Experiments}
\label{sec:experiments}

In this paper, we propose to leverage a large offline experience dataset for efficient demonstration-guided RL. We aim to answer the following questions: (1)~Can the use of task-agnostic prior experience improve the efficiency of \emph{demonstration-guided} RL? (2)~Does the reuse of pre-trained skills reduce the number of required target-specific demonstrations? (3)~In what scenarios does the combination of prior experience and demonstrations lead to the largest efficiency gains?

\subsection{Experimental Setup and Comparisons}

\begin{wrapfigure}{r}{0.5\textwidth}
    \centering
    \vspace{-0.5cm}
    \includegraphics[width=1\linewidth]{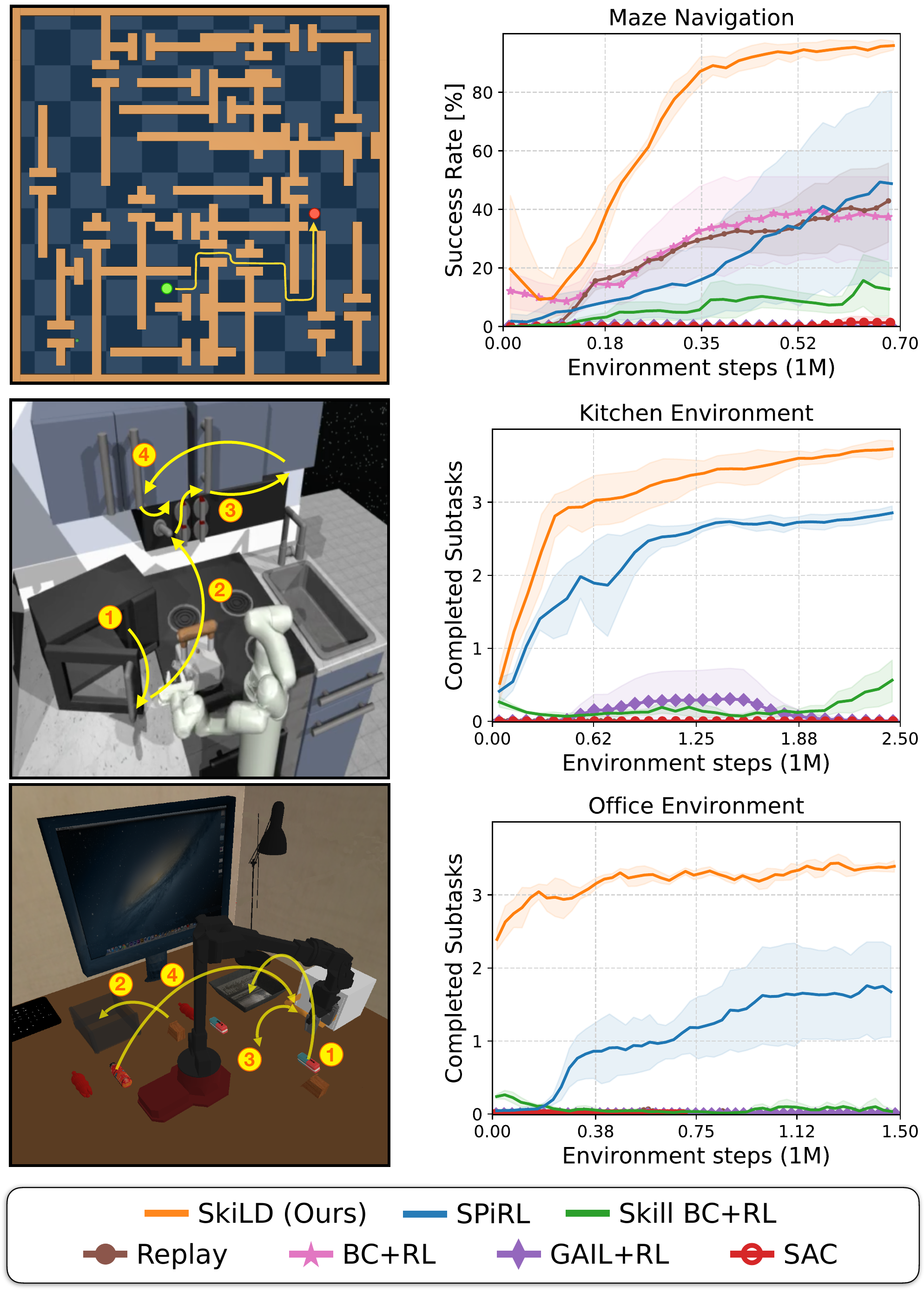}
    \vspace{-0.5cm}
    \caption{\textbf{Left}: Test environments, \textbf{top to bottom}: 2D maze navigation, robotic kitchen manipulation and robotic office cleaning. %
    \textbf{Right}: Target task performance vs environment steps. By using task-agnostic experience, our approach more efficiently leverages the demonstrations than prior demonstration-guided RL approaches across all tasks. The comparison to SPiRL shows that demonstrations improve efficiency even if the agent has access to large amounts of prior experience.%
    }
    \vspace{-0.75cm}
    \label{fig:env_overview_quant_results}
\end{wrapfigure}

To evaluate whether our method \skild~can efficiently use task-agnostic data, we compare it to prior demonstration-guided RL approaches on three complex, long-horizon tasks: a 2D maze navigation task, a robotic kitchen manipulation task and a robotic office cleaning task (see Figure~\ref{fig:env_overview_quant_results}, left).

\paragraph{Maze Navigation.} We adapt the maze navigation task from \citet{pertsch2020spirl} and increase task complexity by adding randomness to the agent's initial position. The agent needs to navigate through the maze to a fixed goal position using planar velocity commands. It only receives a binary reward upon reaching the goal. This environment is challenging for prior demonstration-guided RL approaches since demonstrations of the task span hundreds of time steps, making step-by-step imitation of primitive actions inefficient. %
We collect a task-agnostic offline experience dataset with \SI{3000}{} sequences using a motion planner to find paths between randomly sampled start-goal pairs. This data can be used to extract relevant short-horizon skills like navigating hallways or passing through narrow doors. For the target task we sample an unseen start-goal pair and collect \SI{5}{} demonstrations for reaching the goal position from states sampled around the start position.

\paragraph{Robot Kitchen Environment.} We use the environment of \citet{gupta2019relay} in which a 7DOF robot arm needs to perform a sequence of four subtasks, such as opening the microwave or switching on the light, in the correct order. The agent receives a binary reward upon completion of each consecutive subtask. In addition to the long task horizon, this environment requires precise control of a high-DOF manipulator, testing the scalability of our approach. %
We use \SI{603}{} teleoperated sequences performing various subtask combinations (from~\citet{gupta2019relay}) as our task-agnostic experience dataset $\mathcal{D}$ and separate a set of \SI{20}{} demonstrations for one particular sequence of subtasks, which we define as our target task (see Figure~\ref{fig:env_overview_quant_results}, middle). %

\paragraph{Robot Office Environment.} A 5 DOF robot arm needs to clean an office environment by placing objects in their target bins or putting them in a drawer. It receives binary rewards for the completion of each subtask. In addition to the challenges of the kitchen environment, this task tests the ability of our approach to learn long-horizon behaviors with freely manipulatable objects. %
We collect \SI{2400}{} training trajectories by placing the objects at randomized positions in the environment and performing random subtasks using scripted policies. We also collect \SI{50}{} demonstrations for the unseen target task with new object locations and subtask sequence.

We compare our approach to multiple prior demonstration-guided RL approaches that represent the different classes of existing algorithms introduced in Section~\ref{sec:related_work}. %
In contrast to \skild, these approaches are not designed to leverage task-agnostic prior experience:
\textbf{BC + RL} initializes a policy with behavioral cloning of the demonstrations, then continues to apply BC loss while finetuning the policy with Soft Actor-Critic (SAC, \citep{haarnoja2018sac}), representative of~\citep{rajeswaran2018learning,nair2018overcoming}.
\textbf{GAIL + RL}~\cite{zhu2018reinforcement} combines rewards from the environment and adversarial imitation learning (GAIL, \cite{ho2016generative}), and optimizes the policy using PPO~\cite{schulman2017proximal}.
\textbf{Demo Replay} initializes the replay buffer of an SAC agent with the demonstrations and uses them with prioritized replay during updates, representative of~\citep{vecerik2017leveraging}. 
We also compare our approach to RL-only methods to show the benefit of using demonstration data:
\textbf{SAC} \cite{haarnoja2018sac} is a state-of-the-art model-free RL algorithm, it neither uses offline experience nor demonstrations. 
\textbf{SPiRL} \cite{pertsch2020spirl} extracts skills from task-agnostic experience and performs prior-guided RL on the target task (see Section~\ref{sec:prelim})\footnote{We train SPiRL with the closed-loop policy representation from Section~\ref{sec:subgoal_model} for fair comparison and better performance. For an empirical comparison of open and closed-loop skill representations in SPiRL, see Section~\ref{sec:skill_rep_comparison}.}. 
Finally, we compare to a baseline that combines skills learned from task-agnostic data with target task demonstrations:
\textbf{Skill BC+RL} encodes the demonstrations with the pre-trained skill encoder and runs BC for the high-level skill policy, then finetunes on the target task using SAC. For further details on the environments, data collection, and implementation, see appendix Section~\ref{sec:exp_details}.

\subsection{Demonstration-Guided RL with Learned Skills}
\label{sec:exp_main}

\begin{figure}[t]
    \centering
    \includegraphics[width=1\linewidth]{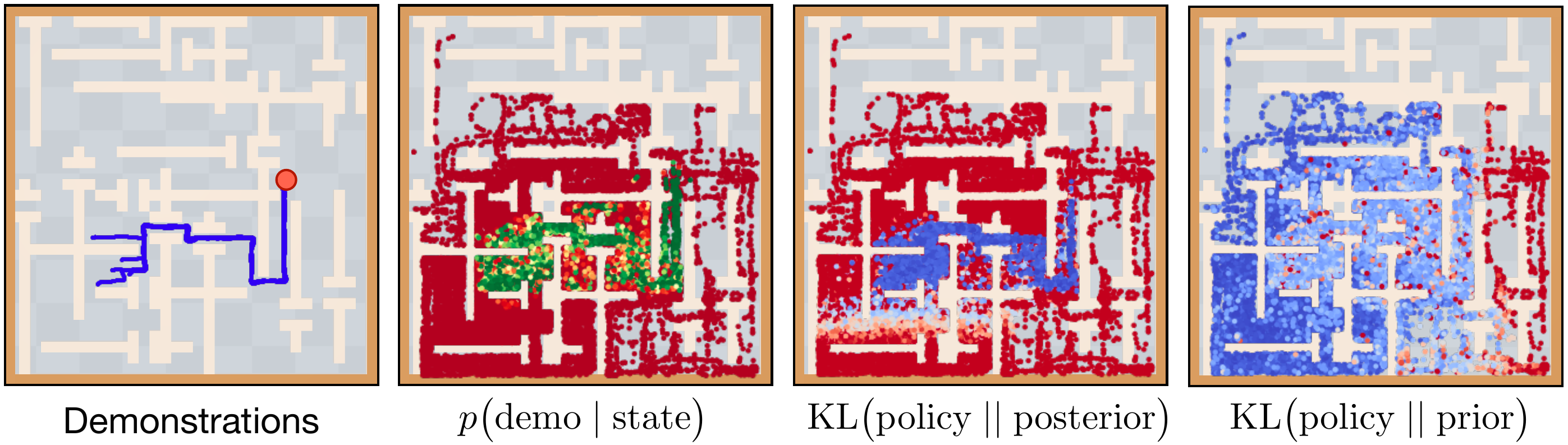}
    \caption{Visualization of our approach on the maze navigation task (%
    visualization states collected by rolling out the skill prior). \textbf{Left}: the given demonstration trajectories; \textbf{Middle left}: output of the demonstration discriminator $D(s)$ (the \textcolor[HTML]{00A86B}{\textbf{greener}}, the higher the predicted probability of a state to be within demonstration support, \textcolor[HTML]{C4051D}{\textbf{red}} indicates low probability). \textbf{Middle right}: policy divergences to the skill posterior and \textbf{Right}: divergence to the skill prior (\textcolor[HTML]{0F52BA}{\textbf{blue}} indicates small and \textcolor[HTML]{C4051D}{\textbf{red}} high divergence). 
    The discriminator accurately infers the demonstration support, the policy successfully follows the skill posterior only within the demonstration support and the skill prior otherwise.%
    }
    \label{fig:maze_vis}
\end{figure}

\paragraph{Maze Navigation.} We compare the downstream task performance of the tested methods on the maze navigation task in Figure~\ref{fig:env_overview_quant_results}~(right). Prior approaches for demonstration-guided RL struggle to learn the task since task-rewards are sparse and only five demonstrations are provided. With such small coverage, behavioral cloning of the demonstrations' primitive actions leads to brittle policies which are hard to finetune (for an analysis of the influence of the number of demonstrations, see Section~\ref{sec:ablations}). The Replay agent improves over SAC without demonstrations and partly succeeds at the task, but learning is slow. The GAIL+RL approach is able to follow part of the demonstrated behavior, but fails to reach the final goal and as a result does not receive the sparse environment reward (see Figure~\ref{fig:maze_gail_quali} for qualitative results). SPiRL and Skill BC+RL, in contrast, are able to leverage offline, task-agnostic experience to learn to occasionally solve the task, but require a substantial amount of environment interactions: SPiRL's learned, task-agnostic skill prior and Skill BC+RL's uniform skill prior during SAC finetuning encourage the policy to try many skills before converging to the ones that solve the downstream task\footnote{Performance of SPiRL differs from \citet{pertsch2020spirl} due to increased task complexity, see Section~\ref{sec:env_details}.}. In contrast, our approach \skild~leverages the task-specific skill posterior to quickly explore the relevant skills, leading to significant efficiency gains (see Figure~\ref{fig:exp_comparison} for a comparison of the exploration of \skild~vs. SPiRL).

We qualitatively analyze our approach in Figure~\ref{fig:maze_vis}: we visualize the output of the discriminator $D(s)$, and the divergences between policy and skill prior and posterior. 
The discriminator accurately estimates the demonstration support, providing a good weighting for prior and posterior regularization, as well as a dense reward bonus. The policy successfully minimizes divergence to the task-specific skill posterior only within the demonstration support and follows the skill prior otherwise.

\paragraph{Robotic Manipulation.} We show the performance comparison on the robotic manipulation tasks in Figure~\ref{fig:env_overview_quant_results}~(right)\footnote{For qualitative robot manipulation videos, see \href{https://clvrai.com/skild}{\url{https://clvrai.com/skild}}.}. %
Both tasks are more challenging since they require precise control of a high-DOF manipulator. We find that approaches for demonstration-guided RL that do not leverage task-agnostic experience struggle to learn either of the tasks since following the demonstrations step-by-step is inefficient and prone to accumulating errors. SPiRL, in contrast, is able to learn meaningful skills from the offline datasets, but struggles to explore the task-relevant skills and therefore learns slowly. Worse yet, the uniform skill prior used in Skill BC+RL's SAC finetuning is even less suited for the target task and prevents the agent from learning the task altogether. Our approach, however, uses the learned skill posterior to guide the chaining of the extracted skills and thereby learns to solve the tasks efficiently, showing how \skild~effectively combines task-agnostic and task-specific data for demonstration-guided RL.

\subsection{Ablation Studies}
\label{sec:ablations}

\begin{wrapfigure}{r}{0.5\textwidth}
    \centering
    \vspace{-0.5cm}
    \includegraphics[width=1\linewidth]{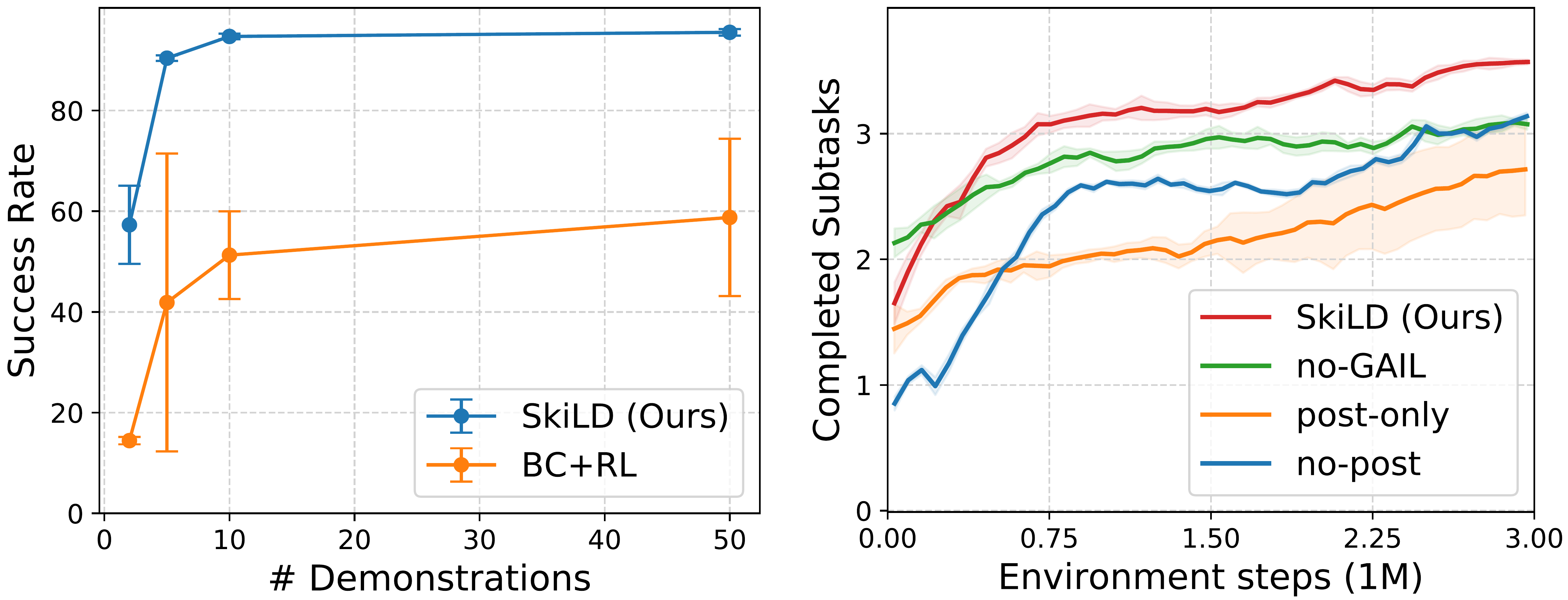}
    \vspace{-0.5cm}
    \caption{Ablation studies. We test the performance of \skild~for different sizes of the demonstration dataset $\vert \mathcal{D}_\text{demo} \vert$ on the maze navigation task (\textbf{left}) and ablate the components of our objective on the kitchen manipulation task (\textbf{right}).
    }
    \vspace{-0.5cm}
    \label{fig:ablation_results}
\end{wrapfigure}

In Figure~\ref{fig:ablation_results} (left) we test the robustness of our approach to the \textbf{number of demonstrations} in the maze navigation task and compare to BC+RL, which we found to work best across different demonstration set sizes. Both approaches benefit from more demonstrations, but our approach is able to learn with much fewer demonstrations by using prior experience. While BC+RL learns each low-level action from the demonstrations, \skild~merely learns to recombine skills it has already mastered using the offline data, thus requiring less dense supervision and fewer demonstrations. %
We also ablate the \textbf{components of our RL objective} on the kitchen task in Figure~\ref{fig:ablation_results} (right). Removing the discriminator reward bonus ("\textit{no-GAIL}") slows convergence since the agent lacks a dense reward signal. Naively replacing the skill prior in the SPiRL objective of Equation~\ref{eq:spirl_objective} with the learned skill \emph{posterior} ("\textit{post-only}") fails since the agent follows the skill posterior outside its support. Removing the skill posterior and optimizing a discriminator bonus augmented reward using SPiRL ("\textit{no-post}") fails because the agent cannot efficiently explore the rich skill space. 
Finally, we show the efficacy of our approach in the pure imitation setting, without environment rewards, in appendix, Section~\ref{sec:skill_imitation}.

\subsection{Data Alignment Analysis}
\label{sec:data_alignment}

\begin{wrapfigure}{r}{0.4\linewidth}
    \vspace{-0.5cm}
    \centering
    \includegraphics[width=1\linewidth]{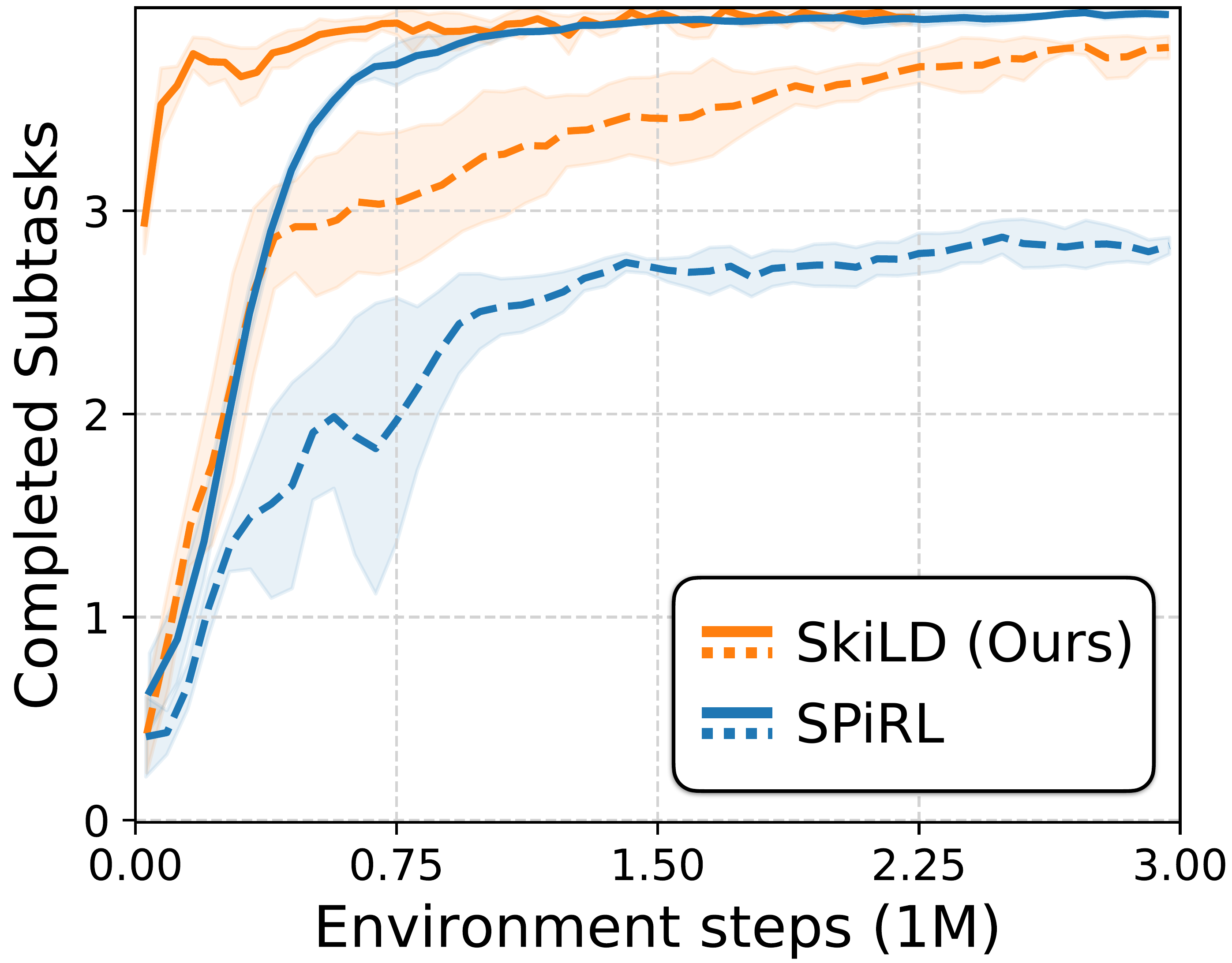}
    \vspace{-0.5cm}
    \caption{Analysis of data vs. task alignment. The benefit of using demonstrations \emph{in addition} to prior experience diminishes if the prior experience is closely aligned with the target task (\textbf{solid}), but gains are high when data and task are not well-aligned (\textbf{dashed}).
    }
    \vspace{-0.5cm}
    \label{fig:data_align_results}
\end{wrapfigure}

We aim to analyze in what scenarios the use of demonstrations \emph{in addition} to task-agnostic experience is most beneficial. In particular, we evaluate how the alignment between the distribution of observed behaviors in the task-agnostic dataset and the target behaviors influences learning efficiency. We choose two different target tasks in the kitchen environment, one with good and one with bad alignment between the behavior distributions, and compare our method, which uses demonstrations, to SPiRL, which only relies on the task-agnostic data\footnote{For a detailed analysis of the behavior distributions in the kitchen dataset and the chosen tasks, see Section~\ref{sec:data_bias_analysis}.}.

In the \textbf{well-aligned} case (Figure~\ref{fig:data_align_results}, solid lines), we find that both approaches learn the task efficiently. Since the skill prior encourages effective exploration on the downstream task, the benefit of the additional demonstrations leveraged in our method is marginal. In contrast, if task-agnostic data and downstream task are \textbf{not well-aligned} (Figure~\ref{fig:data_align_results}, dashed), SPiRL struggles to learn the task since it cannot maximize task reward and minimize divergence from the mis-aligned skill prior at the same time. Our approach learns more reliably by encouraging the policy to reach demonstration-like states and then follow the skill posterior, which by design is well-aligned with the target task. Thus, the gains from our approach are largest when it (a)~focuses a skill prior that explores a too wide range of skills on task-relevant skills or (b)~compensates for mis-alignment between task-agnostic data and target task. %

\section{Conclusion}
\label{sec:conclusion}

We have proposed \skild, a novel approach for demonstration-guided RL that is able to jointly leverage task-agnostic experience datasets and task-specific demonstrations for accelerated learning of unseen tasks. %
In three challenging environments our approach learns unseen tasks more efficiently than both, prior demonstration-guided RL approaches that are not able to leverage task-agnostic experience, as well as skill-based RL methods that cannot effectively incorporate demonstrations.

\clearpage
\bibliography{bibref_definitions_long,bibtex}

\begin{thebibliography}{44}
\providecommand{\natexlab}[1]{#1}
\providecommand{\url}[1]{\texttt{#1}}
\expandafter\ifx\csname urlstyle\endcsname\relax
  \providecommand{\doi}[1]{doi: #1}\else
  \providecommand{\doi}{doi: \begingroup \urlstyle{rm}\Url}\fi

\bibitem[Bekkering et~al.(2000)Bekkering, Wohlschlager, and
  Gattis]{bekkering2000imitation}
H.~Bekkering, A.~Wohlschlager, and M.~Gattis.
\newblock Imitation of gestures in children is goal-directed.
\newblock \emph{The Quarterly Journal of Experimental Psychology: Section A},
  53\penalty0 (1):\penalty0 153--164, 2000.

\bibitem[Al-Abood et~al.(2001)Al-Abood, Davids, and Bennett]{al2001specificity}
S.~A. Al-Abood, K.~Davids, and S.~J. Bennett.
\newblock Specificity of task constraints and effects of visual demonstrations
  and verbal instructions in directing learners' search during skill
  acquisition.
\newblock \emph{Journal of motor behavior}, 33\penalty0 (3):\penalty0 295--305,
  2001.

\bibitem[Hodges et~al.(2007)Hodges, Williams, Hayes, and
  Breslin]{hodges2007modelled}
N.~J. Hodges, A.~M. Williams, S.~J. Hayes, and G.~Breslin.
\newblock What is modelled during observational learning?
\newblock \emph{Journal of sports sciences}, 25\penalty0 (5):\penalty0
  531--545, 2007.

\bibitem[Vecerik et~al.(2017)Vecerik, Hester, Scholz, Wang, Pietquin, Piot,
  Heess, Roth{\"o}rl, Lampe, and Riedmiller]{vecerik2017leveraging}
M.~Vecerik, T.~Hester, J.~Scholz, F.~Wang, O.~Pietquin, B.~Piot, N.~Heess,
  T.~Roth{\"o}rl, T.~Lampe, and M.~Riedmiller.
\newblock Leveraging demonstrations for deep reinforcement learning on robotics
  problems with sparse rewards.
\newblock \emph{arXiv preprint arXiv:1707.08817}, 2017.

\bibitem[Hester et~al.(2018)Hester, Vecerik, Pietquin, Lanctot, Schaul, Piot,
  Horgan, Quan, Sendonaris, Osband, et~al.]{hester2018deep}
T.~Hester, M.~Vecerik, O.~Pietquin, M.~Lanctot, T.~Schaul, B.~Piot, D.~Horgan,
  J.~Quan, A.~Sendonaris, I.~Osband, et~al.
\newblock Deep q-learning from demonstrations.
\newblock In \emph{AAAI}, 2018.

\bibitem[Rajeswaran et~al.(2018)Rajeswaran, Kumar, Gupta, Vezzani, Schulman,
  Todorov, and Levine]{rajeswaran2018learning}
A.~Rajeswaran, V.~Kumar, A.~Gupta, G.~Vezzani, J.~Schulman, E.~Todorov, and
  S.~Levine.
\newblock Learning complex dexterous manipulation with deep reinforcement
  learning and demonstrations.
\newblock In \emph{Robotics: Science and Systems}, 2018.

\bibitem[Nair et~al.(2018)Nair, McGrew, Andrychowicz, Zaremba, and
  Abbeel]{nair2018overcoming}
A.~Nair, B.~McGrew, M.~Andrychowicz, W.~Zaremba, and P.~Abbeel.
\newblock Overcoming exploration in reinforcement learning with demonstrations.
\newblock In \emph{2018 IEEE International Conference on Robotics and
  Automation (ICRA)}, pages 6292--6299. IEEE, 2018.

\bibitem[Zhu et~al.(2018)Zhu, Wang, Merel, Rusu, Erez, Cabi, Tunyasuvunakool,
  Kram{\'a}r, Hadsell, de~Freitas, and Heess]{zhu2018reinforcement}
Y.~Zhu, Z.~Wang, J.~Merel, A.~Rusu, T.~Erez, S.~Cabi, S.~Tunyasuvunakool,
  J.~Kram{\'a}r, R.~Hadsell, N.~de~Freitas, and N.~Heess.
\newblock Reinforcement and imitation learning for diverse visuomotor skills.
\newblock In \emph{Robotics: Science and Systems}, 2018.

\bibitem[Ross et~al.(2011)Ross, Gordon, and Bagnell]{ross2011dagger}
S.~Ross, G.~Gordon, and D.~Bagnell.
\newblock A reduction of imitation learning and structured prediction to
  no-regret online learning.
\newblock In \emph{International Conference on Artificial Intelligence and
  Statistics}, pages 627--635, 2011.

\bibitem[Argall et~al.(2009)Argall, Chernova, Veloso, and
  Browning]{argall2009survey}
B.~D. Argall, S.~Chernova, M.~Veloso, and B.~Browning.
\newblock A survey of robot learning from demonstration.
\newblock \emph{Robotics and autonomous systems}, 57\penalty0 (5):\penalty0
  469--483, 2009.

\bibitem[Pomerleau(1989)]{pomerleau1989alvinn}
D.~A. Pomerleau.
\newblock Alvinn: An autonomous land vehicle in a neural network.
\newblock In \emph{Proceedings of Neural Information Processing Systems
  (NeurIPS)}, pages 305--313, 1989.

\bibitem[Abbeel and Ng(2004)]{abbeel2004apprenticeship}
P.~Abbeel and A.~Y. Ng.
\newblock Apprenticeship learning via inverse reinforcement learning.
\newblock In \emph{ICML}, 2004.

\bibitem[Ho and Ermon(2016)]{ho2016generative}
J.~Ho and S.~Ermon.
\newblock Generative adversarial imitation learning.
\newblock \emph{NeurIPS}, 2016.

\bibitem[Peng et~al.(2018)Peng, Abbeel, Levine, and van~de
  Panne]{peng2018deepmimic}
X.~B. Peng, P.~Abbeel, S.~Levine, and M.~van~de Panne.
\newblock Deepmimic: Example-guided deep reinforcement learning of
  physics-based character skills.
\newblock \emph{ACM Transactions on Graphics (TOG)}, 37\penalty0 (4):\penalty0
  1--14, 2018.

\bibitem[Merel et~al.(2017)Merel, Tassa, TB, Srinivasan, Lemmon, Wang, Wayne,
  and Heess]{merel2017learning}
J.~Merel, Y.~Tassa, D.~TB, S.~Srinivasan, J.~Lemmon, Z.~Wang, G.~Wayne, and
  N.~Heess.
\newblock Learning human behaviors from motion capture by adversarial
  imitation.
\newblock \emph{arXiv preprint arXiv:1707.02201}, 2017.

\bibitem[Pertsch et~al.(2020)Pertsch, Lee, and Lim]{pertsch2020spirl}
K.~Pertsch, Y.~Lee, and J.~J. Lim.
\newblock Accelerating reinforcement learning with learned skill priors.
\newblock In \emph{Conference on Robot Learning (CoRL)}, 2020.

\bibitem[Siegel et~al.(2020)Siegel, Springenberg, Berkenkamp, Abdolmaleki,
  Neunert, Lampe, Hafner, and Riedmiller]{siegel2020keep}
N.~Y. Siegel, J.~T. Springenberg, F.~Berkenkamp, A.~Abdolmaleki, M.~Neunert,
  T.~Lampe, R.~Hafner, and M.~Riedmiller.
\newblock Keep doing what worked: Behavioral modelling priors for offline
  reinforcement learning.
\newblock \emph{ICLR}, 2020.

\bibitem[Nair et~al.(2020)Nair, Dalal, Gupta, and Levine]{nair2020accelerating}
A.~Nair, M.~Dalal, A.~Gupta, and S.~Levine.
\newblock Accelerating online reinforcement learning with offline datasets.
\newblock \emph{arXiv preprint arXiv:2006.09359}, 2020.

\bibitem[Ajay et~al.(2020)Ajay, Kumar, Agrawal, Levine, and
  Nachum]{ajay2020opal}
A.~Ajay, A.~Kumar, P.~Agrawal, S.~Levine, and O.~Nachum.
\newblock Opal: Offline primitive discovery for accelerating offline
  reinforcement learning.
\newblock \emph{arXiv preprint arXiv:2010.13611}, 2020.

\bibitem[Singh et~al.(2021)Singh, Liu, Zhou, Yu, Rhinehart, and
  Levine]{singh2020parrot}
A.~Singh, H.~Liu, G.~Zhou, A.~Yu, N.~Rhinehart, and S.~Levine.
\newblock Parrot: Data-driven behavioral priors for reinforcement learning.
\newblock \emph{ICLR}, 2021.

\bibitem[Singh et~al.(2020)Singh, Yu, Yang, Zhang, Kumar, and
  Levine]{singh2020cog}
A.~Singh, A.~Yu, J.~Yang, J.~Zhang, A.~Kumar, and S.~Levine.
\newblock Cog: Connecting new skills to past experience with offline
  reinforcement learning.
\newblock \emph{CoRL}, 2020.

\bibitem[Hausman et~al.(2018)Hausman, Springenberg, Wang, Heess, and
  Riedmiller]{hausman2018learning}
K.~Hausman, J.~T. Springenberg, Z.~Wang, N.~Heess, and M.~Riedmiller.
\newblock Learning an embedding space for transferable robot skills.
\newblock In \emph{ICLR}, 2018.

\bibitem[Sharma et~al.(2020)Sharma, Gu, Levine, Kumar, and
  Hausman]{sharma2019dynamics}
A.~Sharma, S.~Gu, S.~Levine, V.~Kumar, and K.~Hausman.
\newblock Dynamics-aware unsupervised discovery of skills.
\newblock \emph{ICLR}, 2020.

\bibitem[Gupta et~al.(2019)Gupta, Kumar, Lynch, Levine, and
  Hausman]{gupta2019relay}
A.~Gupta, V.~Kumar, C.~Lynch, S.~Levine, and K.~Hausman.
\newblock Relay policy learning: Solving long-horizon tasks via imitation and
  reinforcement learning.
\newblock \emph{CoRL}, 2019.

\bibitem[Mandlekar et~al.(2018)Mandlekar, Zhu, Garg, Booher, Spero, Tung, Gao,
  Emmons, Gupta, Orbay, Savarese, and Fei-Fei]{mandlekar2018roboturk}
A.~Mandlekar, Y.~Zhu, A.~Garg, J.~Booher, M.~Spero, A.~Tung, J.~Gao, J.~Emmons,
  A.~Gupta, E.~Orbay, S.~Savarese, and L.~Fei-Fei.
\newblock Roboturk: A crowdsourcing platform for robotic skill learning through
  imitation.
\newblock In \emph{CoRL}, 2018.

\bibitem[Lynch et~al.(2020)Lynch, Khansari, Xiao, Kumar, Tompson, Levine, and
  Sermanet]{lynch2020learning}
C.~Lynch, M.~Khansari, T.~Xiao, V.~Kumar, J.~Tompson, S.~Levine, and
  P.~Sermanet.
\newblock Learning latent plans from play.
\newblock In \emph{CoRL}, 2020.

\bibitem[Merel et~al.(2019)Merel, Hasenclever, Galashov, Ahuja, Pham, Wayne,
  Teh, and Heess]{merel2018neural}
J.~Merel, L.~Hasenclever, A.~Galashov, A.~Ahuja, V.~Pham, G.~Wayne, Y.~W. Teh,
  and N.~Heess.
\newblock Neural probabilistic motor primitives for humanoid control.
\newblock \emph{ICLR}, 2019.

\bibitem[Kipf et~al.(2019)Kipf, Li, Dai, Zambaldi, Grefenstette, Kohli, and
  Battaglia]{kipf2018compositional}
T.~Kipf, Y.~Li, H.~Dai, V.~Zambaldi, E.~Grefenstette, P.~Kohli, and
  P.~Battaglia.
\newblock Compositional imitation learning: Explaining and executing one task
  at a time.
\newblock \emph{ICML}, 2019.

\bibitem[Merel et~al.(2020)Merel, Tunyasuvunakool, Ahuja, Tassa, Hasenclever,
  Pham, Erez, Wayne, and Heess]{merel2019reusable}
J.~Merel, S.~Tunyasuvunakool, A.~Ahuja, Y.~Tassa, L.~Hasenclever, V.~Pham,
  T.~Erez, G.~Wayne, and N.~Heess.
\newblock Catch \& carry: Reusable neural controllers for vision-guided
  whole-body tasks.
\newblock \emph{ACM. Trans. Graph.}, 2020.

\bibitem[Shankar et~al.(2019)Shankar, Tulsiani, Pinto, and
  Gupta]{shankar2019discovering}
T.~Shankar, S.~Tulsiani, L.~Pinto, and A.~Gupta.
\newblock Discovering motor programs by recomposing demonstrations.
\newblock In \emph{ICLR}, 2019.

\bibitem[Whitney et~al.(2020)Whitney, Agarwal, Cho, and
  Gupta]{whitney2019dynamics}
W.~Whitney, R.~Agarwal, K.~Cho, and A.~Gupta.
\newblock Dynamics-aware embeddings.
\newblock \emph{ICLR}, 2020.

\bibitem[Lee et~al.(2020)Lee, Yang, and Lim]{lee2020learning}
Y.~Lee, J.~Yang, and J.~J. Lim.
\newblock Learning to coordinate manipulation skills via skill behavior
  diversification.
\newblock In \emph{ICLR}, 2020.

\bibitem[Pertsch et~al.(2020)Pertsch, Rybkin, Yang, Zhou, Derpanis, Lim,
  Daniilidis, and Jaegle]{pertsch2020keyin}
K.~Pertsch, O.~Rybkin, J.~Yang, S.~Zhou, K.~Derpanis, J.~Lim, K.~Daniilidis,
  and A.~Jaegle.
\newblock Keyframing the future: Keyframe discovery for visual prediction and
  planning.
\newblock \emph{L4DC}, 2020.

\bibitem[Higgins et~al.(2017)Higgins, Matthey, Pal, Burgess, Glorot, Botvinick,
  Mohamed, and Lerchner]{higgins2017beta}
I.~Higgins, L.~Matthey, A.~Pal, C.~Burgess, X.~Glorot, M.~Botvinick,
  S.~Mohamed, and A.~Lerchner.
\newblock beta-{VAE}: Learning basic visual concepts with a constrained
  variational framework.
\newblock In \emph{ICLR}, 2017.

\bibitem[Levy et~al.(2019)Levy, Konidaris, Platt, and Saenko]{levy2017learning}
A.~Levy, G.~Konidaris, R.~Platt, and K.~Saenko.
\newblock Learning multi-level hierarchies with hindsight.
\newblock \emph{ICLR}, 2019.

\bibitem[Haarnoja et~al.(2018)Haarnoja, Zhou, Hartikainen, Tucker, Ha, Tan,
  Kumar, Zhu, Gupta, Abbeel, et~al.]{haarnoja2018sac_algo_applications}
T.~Haarnoja, A.~Zhou, K.~Hartikainen, G.~Tucker, S.~Ha, J.~Tan, V.~Kumar,
  H.~Zhu, A.~Gupta, P.~Abbeel, et~al.
\newblock Soft actor-critic algorithms and applications.
\newblock \emph{arXiv preprint arXiv:1812.05905}, 2018.

\bibitem[Finn et~al.(2016)Finn, Christiano, Abbeel, and
  Levine]{finn2016connection}
C.~Finn, P.~Christiano, P.~Abbeel, and S.~Levine.
\newblock A connection between generative adversarial networks, inverse
  reinforcement learning, and energy-based models.
\newblock \emph{NeurIPS Workshop on Adversarial Training}, 2016.

\bibitem[Fu et~al.(2018)Fu, Luo, and Levine]{fu2018learning}
J.~Fu, K.~Luo, and S.~Levine.
\newblock Learning robust rewards with adversarial inverse reinforcement
  learning.
\newblock In \emph{ICLR}, 2018.

\bibitem[Kostrikov et~al.(2019)Kostrikov, Agrawal, Dwibedi, Levine, and
  Tompson]{kostrikov2018discriminatoractorcritic}
I.~Kostrikov, K.~K. Agrawal, D.~Dwibedi, S.~Levine, and J.~Tompson.
\newblock Discriminator-actor-critic: Addressing sample inefficiency and reward
  bias in adversarial imitation learning.
\newblock In \emph{ICLR}, 2019.

\bibitem[Haarnoja et~al.(2018)Haarnoja, Zhou, Abbeel, and
  Levine]{haarnoja2018sac}
T.~Haarnoja, A.~Zhou, P.~Abbeel, and S.~Levine.
\newblock Soft actor-critic: Off-policy maximum entropy deep reinforcement
  learning with a stochastic actor.
\newblock \emph{ICML}, 2018.

\bibitem[Schulman et~al.(2017)Schulman, Wolski, Dhariwal, Radford, and
  Klimov]{schulman2017proximal}
J.~Schulman, F.~Wolski, P.~Dhariwal, A.~Radford, and O.~Klimov.
\newblock Proximal policy optimization algorithms.
\newblock \emph{arXiv preprint arXiv:1707.06347}, 2017.

\bibitem[Liu et~al.(2020)Liu, Jiang, He, Chen, Liu, Gao, and Han]{liu2019radam}
L.~Liu, H.~Jiang, P.~He, W.~Chen, X.~Liu, J.~Gao, and J.~Han.
\newblock On the variance of the adaptive learning rate and beyond.
\newblock In \emph{ICLR}, 2020.

\bibitem[Kingma and Ba(2015)]{kingma2014adam}
D.~P. Kingma and J.~Ba.
\newblock Adam: A method for stochastic optimization.
\newblock In \emph{ICLR}, 2015.

\bibitem[Fu et~al.(2020)Fu, Kumar, Nachum, Tucker, and Levine]{fu2020d4rl}
J.~Fu, A.~Kumar, O.~Nachum, G.~Tucker, and S.~Levine.
\newblock D4rl: Datasets for deep data-driven reinforcement learning.
\newblock \emph{arXiv preprint arXiv:2004.07219}, 2020.

\end{thebibliography}

\clearpage

\appendix
\section{Full Algorithm}
\label{sec:algorithm}

\begin{algorithm*}[t]
\caption{\skild~(\textbf{Ski}ll-based \textbf{L}earning with \textbf{D}emonstrations)}
\label{alg:skil}
\begin{algorithmic}[1]
\STATE \textbf{Inputs:} $H$-step reward function $\tilde{r}(s_t, z_t)$, reward weight $\gamma$, discount $\eta$, target divergences $\delta, \delta_q$, learning rates $\lambda_{\pi}, \lambda_Q, \lambda_\alpha$, target update rate $\tau$.
\STATE Initialize replay buffer $\mathcal{D}$, high-level policy $\pi_\theta(z_t\vert s_t)$, critic $Q_\phi(s_t, z_t)$, target network $Q_{\bar{\phi}}(s_t, z_t)$
\FOR{each iteration}

\FOR{every $H$ environment steps}
\STATE $z_t \sim \pi(z_t \vert s_t)$ \COMMENT{sample skill from policy}
\STATE $s_{t^\prime} \sim p(s_{t+H} \vert s_t, z_t)$ \COMMENT{execute skill in environment}
\STATE $\mathcal{D} \leftarrow \mathcal{D} \cup \{s_t, z_t, \tilde{r}(s_t, z_t), s_{t^\prime}\}$ \COMMENT{store transition in replay buffer}
\ENDFOR

\FOR{each gradient step}
\STATE $\textcolor{red}{r_\Sigma = (1 - \gamma) \cdot \tilde{r}(s_t, z_t) + \gamma \cdot \big[\log D(s_t) - \log\big(1 - D(s_t)\big)\big]}$ \COMMENT{compute combined reward}

\STATE $\bar{Q} = \textcolor{red}{r_\Sigma} + \eta \big[ Q_{\bar{\phi}}(s_{t^\prime}, \pi_\theta(z_{t^\prime} \vert s_{t^\prime})) \textcolor{red}{- \big[\alpha_q D_\text{KL}\big(\pi_\theta(z_{t^\prime} \vert s_{t^\prime}), q_\zeta(z_{t^\prime} \vert s_{t^\prime})\big) \cdot D(s_{t^\prime})}$
\STATE \hspace{5cm}$\textcolor{red}{+\;\alpha D_\text{KL}\big(\pi_\theta(z_{t^\prime} \vert s_{t^\prime}), p_\va(z_{t^\prime} \vert s_{t^\prime})\big) \cdot \big(1 - D(s_{t^\prime})\big) \big]}$ \COMMENT{compute Q-target}

\STATE $\theta \leftarrow \theta - \lambda_\pi \nabla_\theta \big[Q_\phi(s_t, \pi_\theta(z_t \vert s_t)) \textcolor{red}{- \big[\alpha_q D_\text{KL}\big(\pi_\theta(z_{t} \vert s_{t}), q_\zeta(z_{t} \vert s_{t})\big) \cdot D(s_{t})}$
\STATE \hspace{5cm}$\textcolor{red}{+\;\alpha D_\text{KL}\big(\pi_\theta(z_{t} \vert s_{t}), p_\va(z_{t} \vert s_{t})\big) \cdot \big(1 - D(s_{t})\big) \big]}$ \COMMENT{update policy weights}
\STATE $\phi \leftarrow \phi - \lambda_Q \nabla_\phi \big[ \frac{1}{2}\big(Q_\phi(s_t, z_t) - \bar{Q} \big)^2 \big]$ \COMMENT{update critic weights}
\STATE $\alpha \leftarrow \alpha - \lambda_\alpha \nabla_\alpha \big[ \alpha \cdot (\textcolor{red}{D_\text{KL}(\pi_\theta(z_t \vert s_t), p_\va(z_t \vert s_t)) - \delta}) \big]$ \COMMENT{update alpha}
\STATE $\textcolor{red}{\alpha_q \leftarrow \alpha_q - \lambda_{\alpha} \nabla_{\alpha_q} \big[ \alpha_q \cdot (D_\text{KL}(\pi_\theta(z_t \vert s_t), q_\zeta(z_t \vert s_t)) - \delta_q}) \big]$ \COMMENT{update alpha-q}
\STATE $\bar{\phi} \leftarrow \tau \phi + (1 - \tau) \bar{\phi}$ \COMMENT{update target network weights}
\ENDFOR

\ENDFOR
\STATE \textbf{return} trained policy $\pi_\theta(z_t \vert s_t)$
\end{algorithmic}
\end{algorithm*}

We detail our full \skild~algorithm for demonstration-guided RL with learned skills in Algorithm~\ref{alg:skil}. It is based on the SPiRL algorithm for RL with learned skills~\cite{pertsch2020spirl} which in turn builds on Soft-Actor Critic~\cite{haarnoja2018sac}, an off-policy model-free RL algorithm. We mark changes of our algorithm with respect to SPiRL and SAC in \textcolor{red}{red} in Algorithm~\ref{alg:skil}.

The hyperparameters $\alpha$ and $\alpha_q$ can either be constant, or they can be automatically tuned using dual gradient descent~\cite{haarnoja2018sac_algo_applications,pertsch2020spirl}. In the latter case, we need to define a set of \emph{target divergences} $\delta, \delta_q$. The parameters $\alpha$ and $\alpha_q$ are then optimized to ensure that the expected divergence between policy and skill prior and posterior distributions is equal to the chosen target divergence (see Algorithm~\ref{alg:skil}).

\section{Implementation and Experimental Details}
\label{sec:exp_details}

\subsection{Implementation Details: Pre-Training}

We introduce our objective for learning the skill inference network $q_\omega(z\vert s,a)$ and low-level skill policy $\pi_\phi(a_t\vert s_t, z)$ in Section~\ref{sec:subgoal_model}. In practice, we instantiate all model components with deep neural networks $Q_\omega, \Pi_\phi$ respectively, and optimize the full model using back-propagation. We also jointly train our skill prior network $P_\va$. We follow the common assumption of Gaussian, unit-variance output distributions for low-level policy actions, leading to the following network loss:
\begin{align*}
    \mathcal{L} = &\underbrace{\prod_{t=0}^{H-2}\norm[\big]{a_t - \Pi_\phi(s_t, z)}^2
    + \beta D_{\text{KL}}\big(Q_\omega(s_{0:H-1}, a_{0:H-2}) \;\vert\vert\; \mathcal{N}(0, I)\big)}_\text{skill representation training}\\
    &+ \underbrace{D_{\text{KL}}\big(\lfloor Q_\omega(s_{0:H-1}, a_{0:H-2}) \rfloor \;\vert\vert\; P_\va(s_0)\big)}_\text{skill prior training}.
\end{align*}

Here $\lfloor \cdot \rfloor$ indicates that we stop gradient flow from the prior training objective into the skill inference network for improved training stability. After training the skill inference network with above objective, we train the skill posterior network $Q_\zeta$ by minimizing KL divergence to the skill inference network's output on trajectories sampled from the demonstration data. We minimize the following objective:
\begin{equation*}
    \mathcal{L}_\text{post} = D_{\text{KL}}\big(\lfloor Q_\omega(s_{0:H-1}, a_{0:H-2}) \rfloor \;\vert\vert\; Q_\zeta(s_0)\big)
\end{equation*}

We use a 1-layer LSTM with 128 hidden units for the inference network and 3-layer MLPs with 128 hidden units in each layer for the low-level policy. We encode skills of horizon \SI{10}{} into \SI{10}{}-dimensional skill representations $z$. Skill prior and posterior networks are implemented as 5-layer MLPs with 128 hidden units per layer. They both parametrize mean and standard deviation of Gaussian output distributions. All networks use batch normalization after every layer and leaky ReLU activation functions. We tune the regularization weight $\beta$ to be \SI{1e-3}{} for the maze and \SI{5e-4}{} for kitchen and office environment.

For the demonstration discriminator $D(s)$ we use a 3-layer MLP with only 32 hidden units per layer to avoid overfitting. It uses a sigmoid activation function on the final layer and leaky ReLU activations otherwise. We train the discriminator with binary cross-entropy loss on samples from task-agnostic and demonstration datasets:
\begin{equation*}
    \mathcal{L}_\text{D} = - \frac{1}{N} \cdot \big[\underbrace{\sum_{i=1}^{N/2} \log D(s_i^d)}_{\text{demonstrations}} + \underbrace{\sum_{j=1}^{N/2} \log\big(1 - D(s_j)\big)}_{\text{task-agnostic data}}\big]
\end{equation*}

We optimize all networks using the RAdam optimizer~\cite{liu2019radam} with parameters $\beta_1 = 0.9$ and $\beta_2= 0.999$, batch size \SI{128}{} and learning rate \SI{1e-3}{}. On a single NVIDIA Titan X GPU we can train the skill representation and skill prior in approximately \SI{5}{} hours, the skill posterior in approximately \SI{3}{} hours and the discriminator in approximately \SI{3}{} hours.

\subsection{Implementation Details: Downstream RL}

The architecture of the policy mirrors the one of the skill prior and posterior networks. The critic is a simple 2-layer MLP with 256 hidden units per layer. The policy outputs the parameters of a Gaussian action distribution while the critic outputs a single $Q$-value estimate. We initialize the policy with the weights of the skill posterior network.

We use the hyperparameters of the standard SAC implementation~\cite{haarnoja2018sac} with batch size 256, replay buffer capacity of \SI{1e6}{} and discount factor $\gamma = 0.99$. We collect \SI{5000}{} warmup rollout steps to initialize the replay buffer before training. We use the Adam optimizer~\cite{kingma2014adam}
with $\beta_1=0.9$, $\beta_2=0.999$ and learning rate \SI{3e-4}{} for updating policy, critic and temperatures $\alpha$ and $\alpha_q$. Analogous to SAC, we train two separate critic networks and compute the $Q$-value as the minimum over both estimates to stabilize training. The corresponding target networks get updated at a rate of $\tau$ = \SI{5e-3}{}. The policy's actions are limited in the range $[-2, 2]$ by a $\tanh$ "squashing function" (see~\citet{haarnoja2018sac}, appendix C).

We use automatic tuning of $\alpha$ and $\alpha_q$ in the maze navigation task and set the target divergences to \SI{1}{} and \SI{10}{} respectively. In the kitchen and office environments we obtained best results by using constant values of $\alpha = \alpha_q = \SI{1e-1}{}$. In all experiments we set $\kappa = \SI{0.9}{}$.

For all RL results we average the results of three independently seeded runs and display mean and standard deviation across seeds.

\subsection{Implementation Details: Comparisons}

\paragraph{BC+RL.} This comparison is representative of demonstration-guided RL approaches that use BC objectives to initialize and regularize the policy during RL~\cite{rajeswaran2018learning,nair2018overcoming}. We pre-train a BC policy on the demonstration dataset and use it to initialize the RL policy. We use SAC to train the policy on the target task. Similar to~\citet{nair2018overcoming} we augment the policy update with a regularization term that minimizes the L2 loss between the predicted mean of the policy's output distribution and the output of the BC pre-trained policy\footnote{We also tried sampling action targets directly from the demonstration replay buffer, but found using a BC policy as target more effective on the tested tasks.}.

\paragraph{Demo Replay.} This comparison is representative of approaches that initialize the replay buffer of an off-policy RL agent with demonstration transitions~\cite{vecerik2017leveraging,hester2018deep}. In practice we use SAC and initialize a second replay buffer with the demonstration transitions. Since the demonstrations do not come with reward, we heuristically set the reward of each demonstration trajectory to be a high value (\SI{100}{} for the maze, \SI{4}{} for the robotic environments) on the final transition and zero everywhere else. During each SAC update, we sample half of the training mini-batch from the normal SAC replay buffer and half from the demonstration replay buffer. All other aspects of SAC remain unchanged.

\subsection{Environment Details}
\label{sec:env_details}

\begin{figure}[t]
    \centering
    \begin{minipage}{.3\textwidth}
        \centering
        \includegraphics[width=\linewidth]{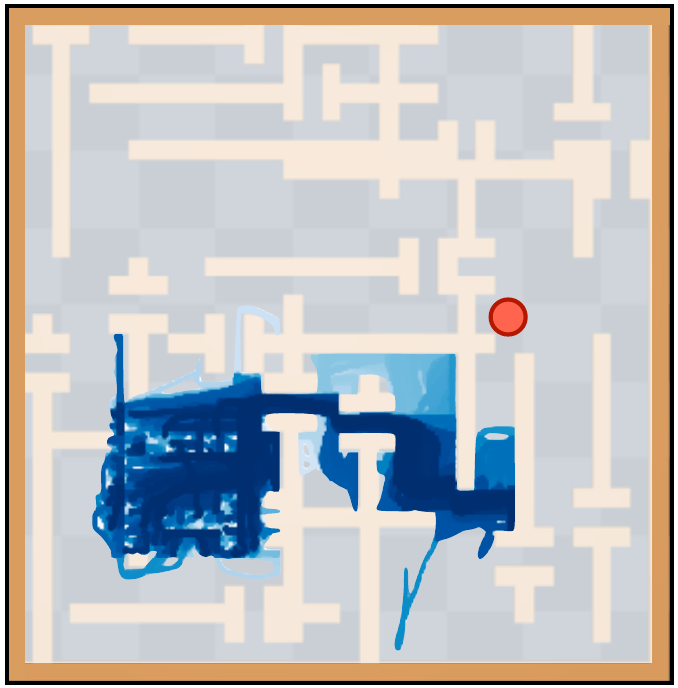}
        \caption{Qualitative results for GAIL+RL on maze navigation. Even though it makes progress towards the goal (\textbf{red}), it fails to ever obtain the sparse goal reaching reward.
        }
        \label{fig:maze_gail_quali}
    \end{minipage}%
    \hspace{1cm}
    \begin{minipage}{.6\textwidth}
        \centering
        \includegraphics[width=\linewidth]{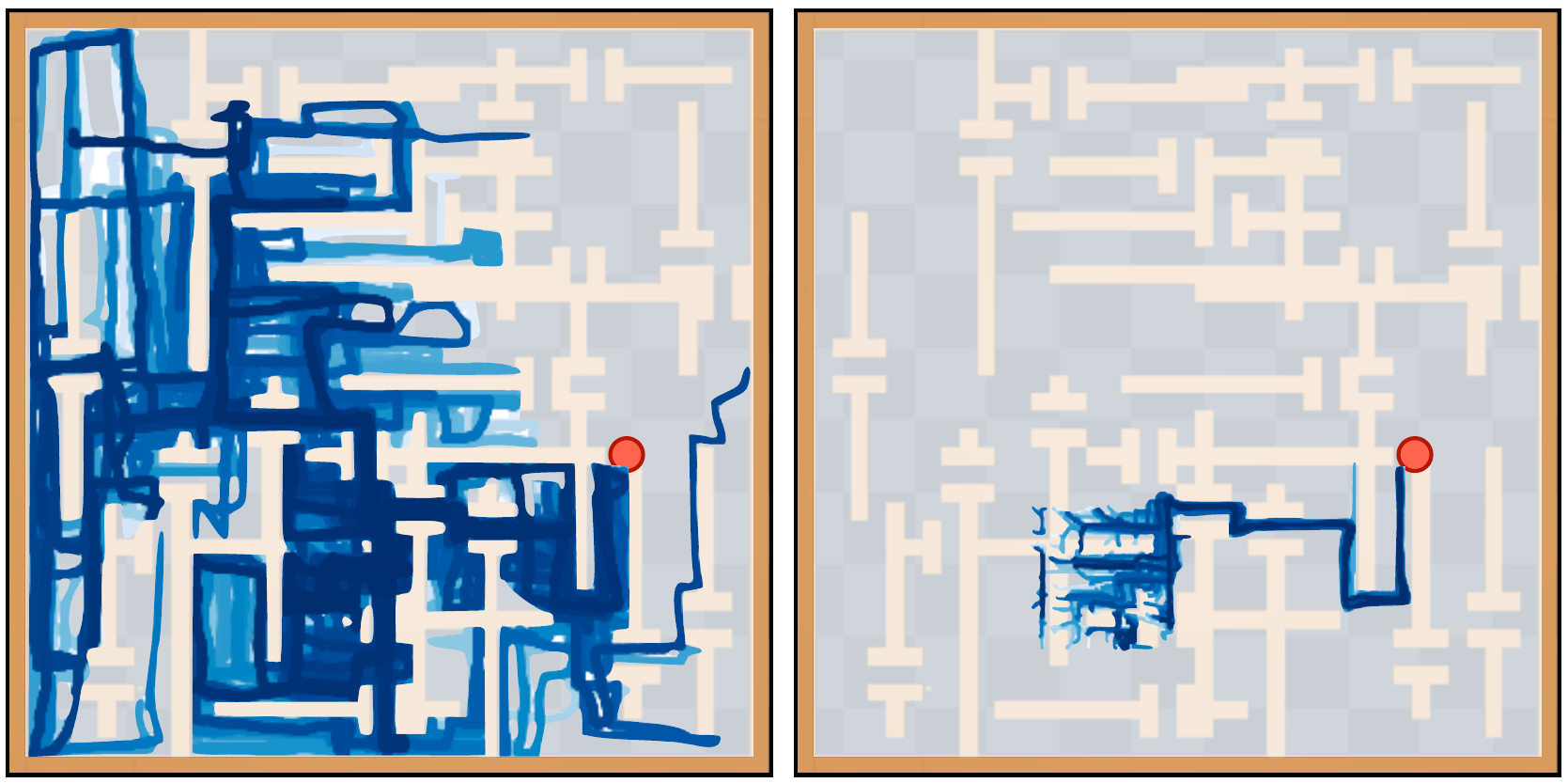}
        \caption{We compare the exploration behavior in the maze. We roll out skills sampled from SPiRL's task-agnostic skill prior (\textbf{left}) and our task-specific skill posterior (\textbf{right}) and find that the latter leads to more targeted exploration towards the goal (red).
        }
        \label{fig:exp_comparison}
    \end{minipage}
\end{figure}

\begin{wrapfigure}{r}{0.5\linewidth}
\vspace{-0.5cm}
    \centering
    \includegraphics[width=\linewidth]{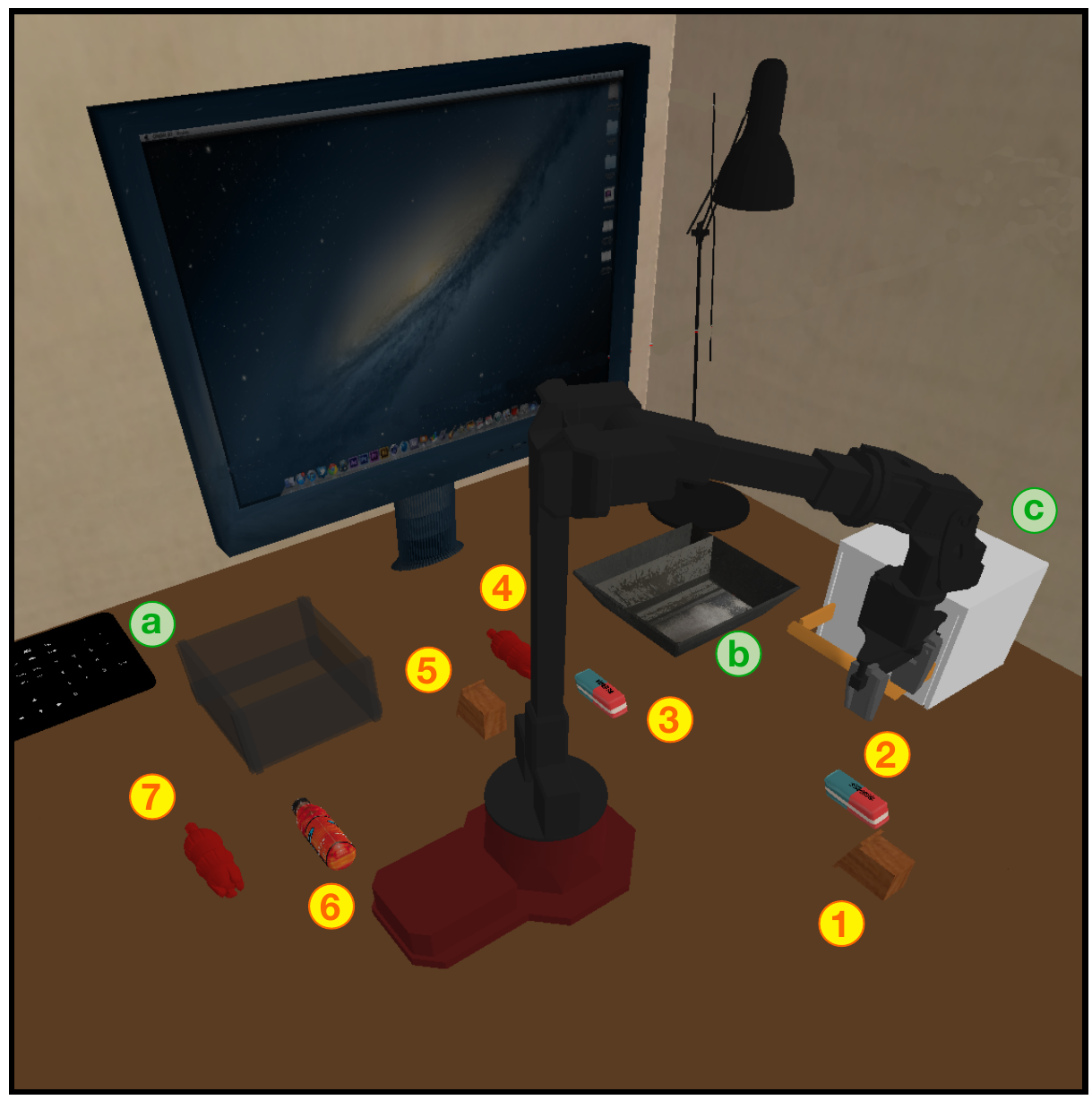}
    \vspace{-0.5cm}
    \caption{Office cleanup task. The robot agent needs to place three randomly sampled objects \textbf{(1-7)} inside randomly sampled containers \textbf{(a-c)}. During task-agnostic data collection we apply random noise to the initial position of the objects.
    }
    \label{fig:office_vis}
    \vspace{-1cm}
\end{wrapfigure}

\paragraph{Maze Navigation.} We adapt the maze navigation task from \citet{pertsch2020spirl} which extends the maze navigation tasks from the D4RL benchmark~\cite{fu2020d4rl}. The starting position is sampled uniformly from a start region and the agent receives a one-time sparse reward of 100 when reaching the fixed goal position, which also ends the episode. The 4D observation space contains 2D position and velocity of the agent. The agent is controlled via 2D velocity commands.

\paragraph{Robot Kitchen Environment.} We use the kitchen environment from~\citet{gupta2019relay}. For solving the target task, the agent needs to execute a fixed sequence of four subtasks by controlling an Emika Franka Panda 7-DOF robot via joint velocity and continuous gripper actuation commands. The 30-dimensional state space contains the robot's joint angles as well as object-specific features that characterize the position of each of the manipulatable objects. We use \SI{20}{} state-action sequences from the dataset of~\citet{gupta2019relay} as demonstrations. Since the dataset does not have large variation \emph{within} the demonstrations for one task, the support of those demonstration is very narrow. We collect a demonstration dataset with widened support by initializing the environment at states along the demonstrations and rolling out a random policy for \SI{10}{} steps. %

\paragraph{Robot Office Environment.} We create a novel office cleanup task in which a 5-DOF WidowX robot needs to place a number of objects into designated containers, requiring the execution of a sequence of pick, place and drawer open and close subtasks (see Figure~\ref{fig:office_vis}). The agent controls position and orientation of the end-effector and a continuous gripper actuation, resulting in a \SI{7}{}-dimensional action space. For simulating the environment we build on the Roboverse framework~\cite{singh2020cog}. During collection of the task-agnostic data we randomly sample a subset of three of the seven objects as well as a random order of target containers and use scripted policies to execute the task. %
We only save successful executions. For the target task we fix object positions and require the agent to place three objects in fixed target containers. The \SI{97}{}-dimensional state space contains the agent's end-effector position and orientation as well as position and orientation of all objects and containers.

\paragraph{Differences to \citet{pertsch2020spirl}.} While both maze navigation and kitchen environment are based on the tasks in \citet{pertsch2020spirl}, we made multiple changes to increase task complexity, resulting in the lower absolute performance of the SPiRL baseline in Figure~\ref{fig:env_overview_quant_results}. For the maze navigation task we added randomness to the starting position and terminate the episode upon reaching the goal position, reducing the max. reward obtainable for successfully solving the task. We also switched to a low-dimensional state representation for simplicity. For the kitchen environment, the task originally used in \citet{gupta2019relay} as well as \citet{pertsch2020spirl} was well aligned with the training data distribution and there were no demonstrations available for this task. In our evaluation we use a different downstream task (see section~\ref{sec:data_bias_analysis}) which is less well-aligned with the training data and therefore harder to learn. This also allows us to use sequences from the dataset of~\citet{gupta2019relay} as demonstrations for this task.

\section{Skill Representation Comparison}
\label{sec:skill_rep_comparison}

\begin{wrapfigure}{r}{0.5\textwidth}
\vspace{-0.5cm}
    \centering
    \includegraphics[width=1\linewidth]{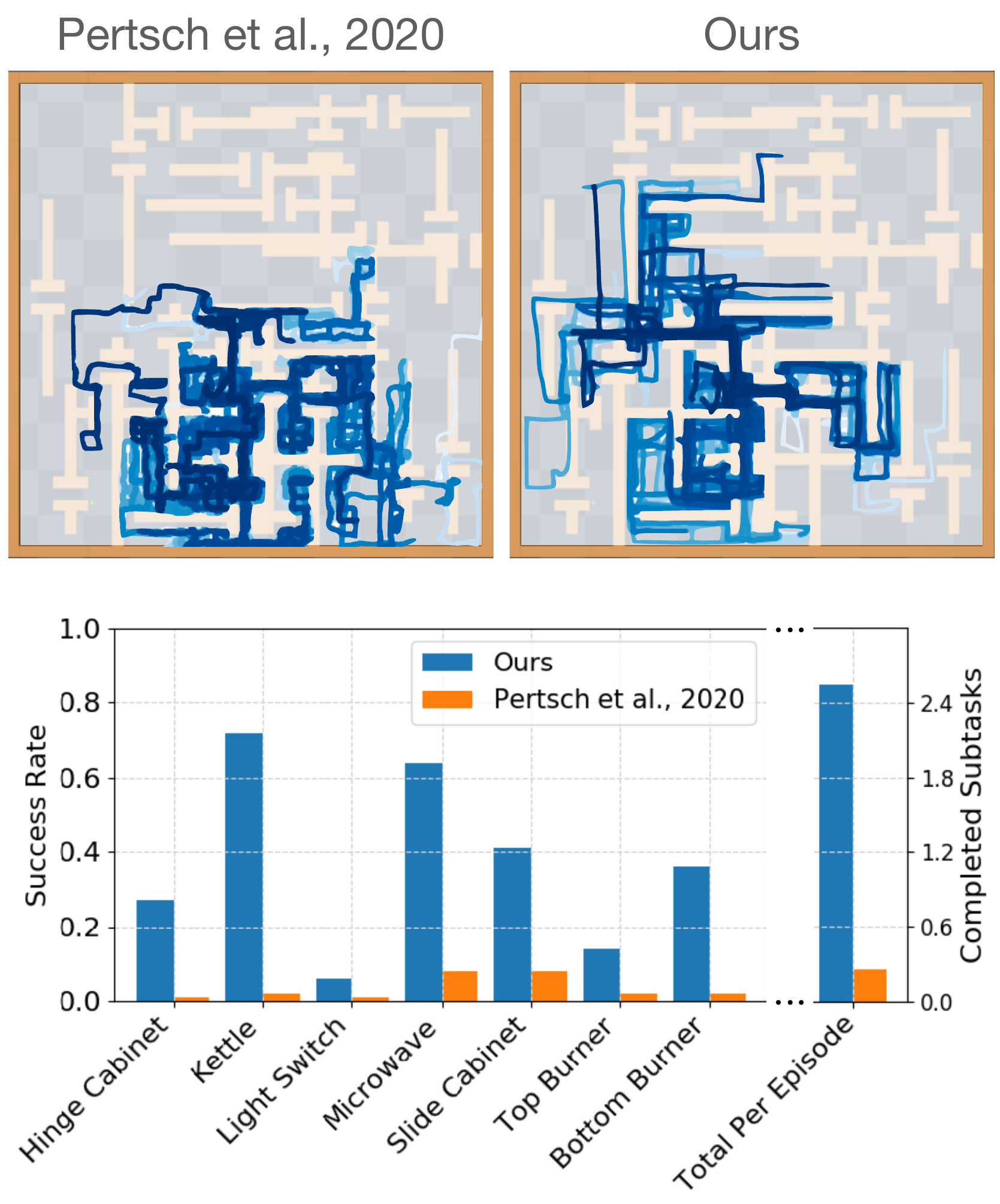}
    \vspace{-0.5cm}
    \caption{Comparison of our closed-loop skill representation with the open-loop representation of~\citet{pertsch2020spirl}. \textbf{Top}: Skill prior rollouts for \SI{100}{k} steps in the maze environment. \textbf{Bottom}: Subtask success rates for prior rollouts in the kitchen environment. 
    }
    \label{fig:skill_rep_results}
    \vspace{-0.5cm}
\end{wrapfigure}

In Section~\ref{sec:subgoal_model} we described our skill representation based on a closed-loop low-level policy as a more powerful alternative to the open-loop action decoder-based representation of~\citet{pertsch2020spirl}. To compare the performance of the two representations we perform rollouts with the learned skill prior: we sample a skill from the prior and rollout the low-level policy for $H$ steps. We repeat this until the episode terminates and visualize the results for multiple episodes in maze and kitchen environment in Figure~\ref{fig:skill_rep_results}.

In Figure~\ref{fig:skill_rep_results} (top) we see that both representations lead to effective exploration in the maze environment. Since the 2D maze navigation task does not require control in high-dimensional action spaces, both skill representations are sufficient to accurately reproduce behaviors observed in the task-agnostic training data.

In contrast, the results on the kitchen environment (Figure~\ref{fig:skill_rep_results}, bottom) show that the closed-loop skill representation is able to more accurately control the high-DOF robotic manipulator and reliably solve multiple subtasks per rollout episode.\footnote{See \href{https://sites.google.com/view/skill-demo-rl}{\url{https://sites.google.com/view/skill-demo-rl}} for skill prior rollout videos with both skill representations in the kitchen environment.} We hypothesize that the closed-loop skill policy is able to learn more robust skills from the task-agnostic training data, particularly in high-dimensional control problems.

\section{Demonstration-Guided RL Comparisons with Task-Agnostic Experience}
\label{sec:demoRL_with_prior}
\begin{figure}[t]
    \centering
    \includegraphics[width=1\linewidth]{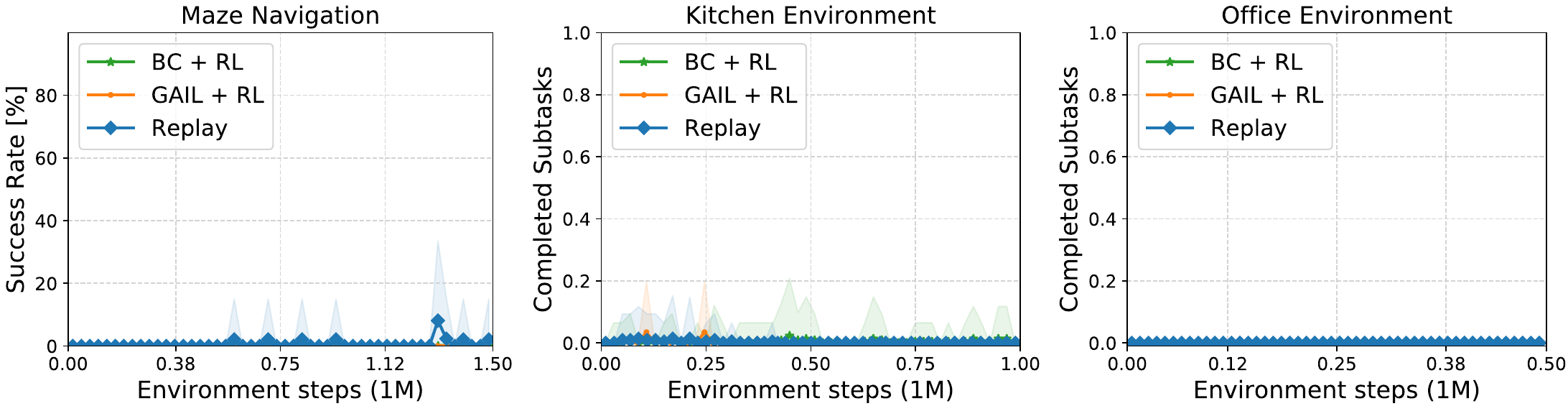}
    \caption{Downstream task performance for prior demonstration-guided RL approaches with combined task-agnostic and task-specific data. All prior approaches are unable to leverage the task-agnostic data, showing a performance decrease when attempting to use it.
    }
    \label{fig:comb_baseline_results}
\end{figure}

In Section~\ref{sec:exp_main} we compared our approach to prior demonstration-guided RL approaches which are not designed to leverage task-agnostic datasets. We applied these prior works in the setting they were designed for: using only task-specific demonstrations of the target task. Here, we conduct experiments in which we run these prior works using the \emph{combined} task-agnostic and task-specific datasets to give them access to the same data that our approach used.

From the results in Figure~\ref{fig:comb_baseline_results} we can see that none of the prior works is able to effectively leverage the additional task-agnostic data. In many cases the performance of the approaches is worse than when only using task-specific data (see Figure~\ref{fig:env_overview_quant_results}). Since prior approaches are not designed to leverage task-agnostic data, applying them in the combined-data setting can hurt learning on the target task. In contrast, our approach can effectively leverage the task-agnostic data for accelerating demonstration-guided RL.

\section{Skill-Based Imitation Learning}
\label{sec:skill_imitation}

\begin{figure}[t]%
    \centering
    \includegraphics[width=1\linewidth]{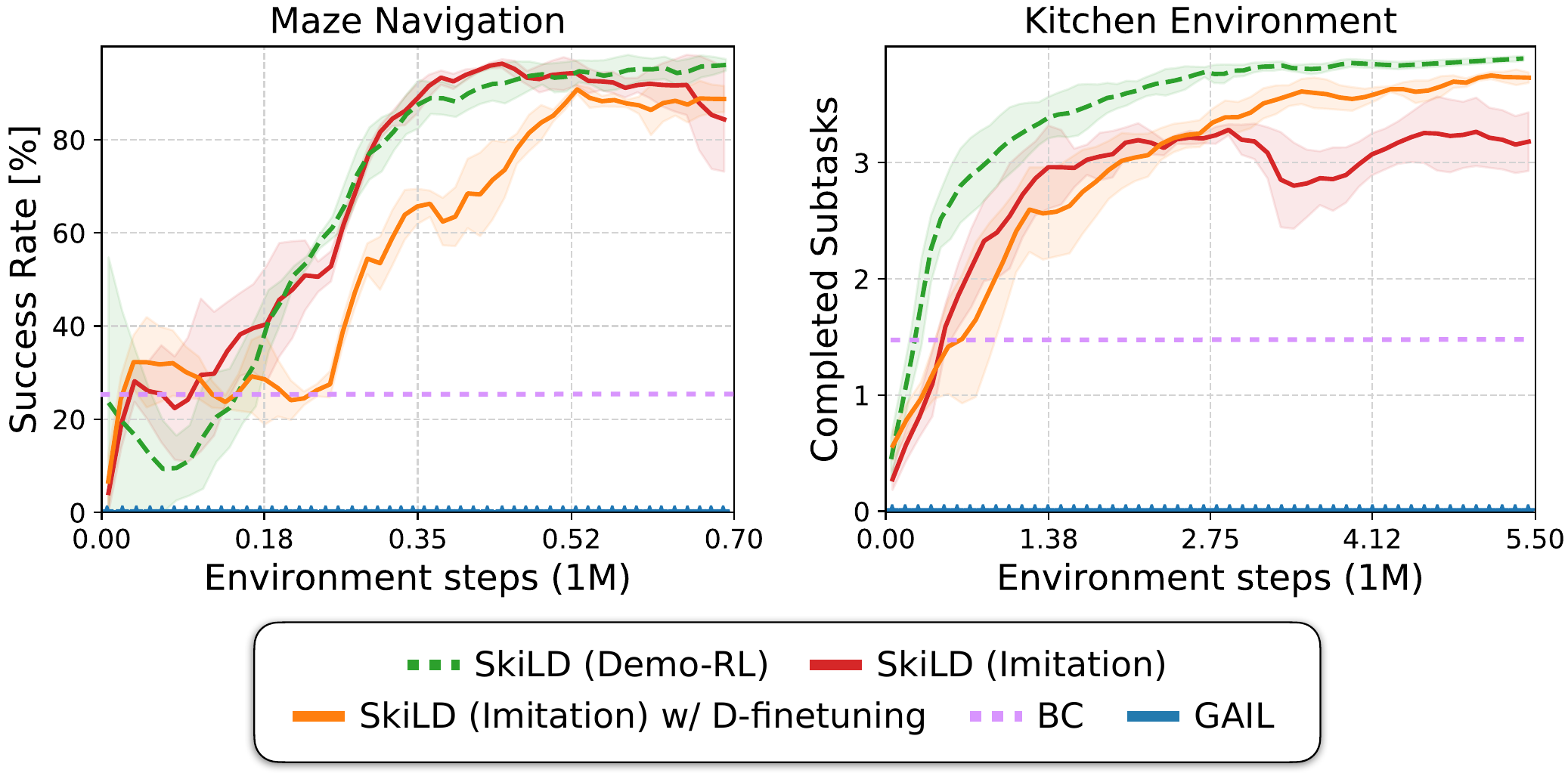}
    \caption{Imitation learning performance on maze navigation and kitchen tasks. Compared to prior imitation learning methods, \skild~can leverage prior experience to enable the imitation of complex, long-horizon behaviors. Finetuning the pre-trained discriminator $D(s)$ further improves performance on more challenging control tasks like in the kitchen environment.
    }
    \label{fig:imitation_results}
\end{figure}

We ablate the influence of the environment reward feedback on the performance of our approach by setting the reward weight $\kappa = 1.0$, thus relying solely on the learned discriminator reward. Our goal is to test whether our approach \skild~is able to leverage task-agnostic experience to improve the performance of pure \emph{imitation learning}, \ie learning to follow demonstrations without environment reward feedback.

We compare \skild~to common approaches for imitation learning: behavioral cloning~(BC, \citet{pomerleau1989alvinn}) and generative adversarial imitation learning~(GAIL, \citet{ho2016generative}). We also experiment with a version of our skill-based imitation learning approach that performs online finetuning of the pre-trained discriminator $D(s)$ using data collected during training of the imitation policy.

We summarize the results of the imitation learning experiments in Figure~\ref{fig:imitation_results}. Learning purely by imitating the demonstrations, without additional reward feedback, is generally slower than demonstration-guided RL on tasks that require more challenging control, like in the kitchen environment, where the pre-trained discriminator does not capture the desired trajectory distribution accurately. Yet, we find that our approach is able to leverage task-agnostic data to effectively imitate complex, long-horizon behaviors while prior imitation learning approaches struggle. Further, online finetuning of the learned discriminator improves imitation learning performance when the pre-trained discriminator is not accurate enough. 

In the maze navigation task the pre-trained discriminator represents the distribution of solution trajectories well, so pure imitation performance is comparable to demonstration-guided RL. We find that finetuning the discriminator on the maze ``sharpens'' the decision boundary of the discriminator, \ie increases its confidence in correctly estimating the demonstration support. Yet, this does not lead to faster overall convergence since the pre-trained discriminator is already sufficiently accurate.

\section{Kitchen Data Analysis}
\label{sec:data_bias_analysis}

\begin{figure*}[t]
    \centering
    \includegraphics[width=1\linewidth]{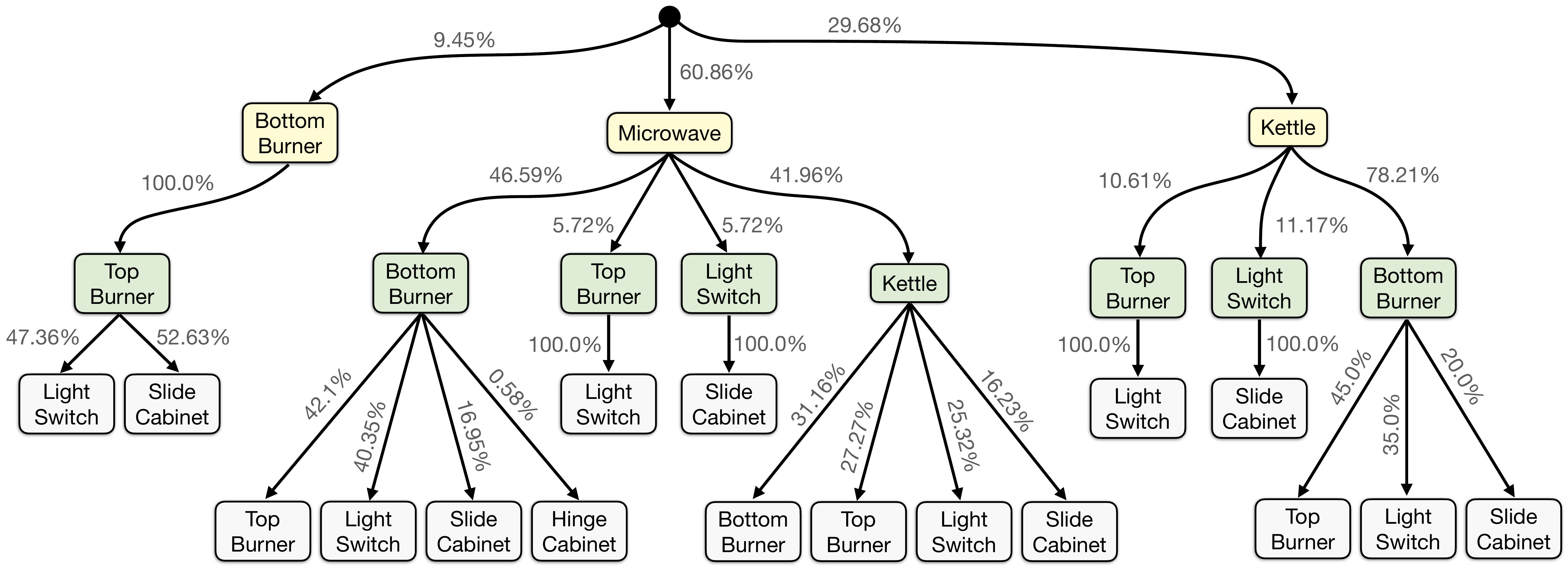}
    \caption{Subtask transition probabilities in the kitchen environment's task-agnostic training dataset from \citet{gupta2019relay}. Each dataset trajectory consists of four consecutive subtasks, of which we display three (\textcolor[HTML]{F4C430}{\textbf{yellow}}: first, \textcolor[HTML]{00A86B}{\textbf{green}}: second, \textcolor[HTML]{5E5E5E}{\textbf{grey}}: third subtask). The transition probability to the fourth subtask is always near \SI{100}{\percent}. In Section~\ref{sec:data_alignment} we test our approach on a target task with good alignment to the task-agnostic data (\textit{Microwave - Kettle - Light Switch - Hinge Cabinet}) and a target task which is mis-aligned to the data (\textit{Microwave - Light Switch - Slide Cabinet - Hinge Cabinet}).
    }
    \label{fig:kitchen_skill_tree}
\end{figure*}

For the kitchen manipulation experiments we use the dataset provided by \citet{gupta2019relay} as task-agnostic pre-training data. It consists of \SI{603}{} teleoperated sequences, each of which shows the completion of four consecutive subtasks. In total there are seven possible subtasks: opening the microwave, moving the kettle, turning on top and bottom burner, flipping the light switch and opening a slide and a hinge cabinet. 

In Figure~\ref{fig:kitchen_skill_tree} we analyze the transition probabilities between subtasks in the task-agnostic dataset. We can see that these transition probabilities are not uniformly distributed, but instead certain transitions are more likely than others, \eg it is much more likely to sample a training trajectory in which the agent first opens the microwave than one in which it starts by turning on the bottom burner. 

In Section~\ref{sec:data_alignment} we test the effect this bias in transition probabilities has on the learning of target tasks. Concretely, we investigate two cases: good alignment between task-agnostic data and target task and mis-alignment between the two. In the former case we choose the target task \textit{Kettle - Bottom Burner - Top Burner - Slide Cabinet}, since the required subtask transitions are likely under the training data distribution. For the mis-aligned case we choose \textit{Microwave - Light Switch - Slide Cabinet - Hinge Cabinet} as target task, since particularly the transition from opening the microwave to flipping the light switch is very unlikely to be observed in the training data.

\end{document}